\documentclass[10pt,journal,compsoc]{IEEEtran}
\usepackage[T1]{fontenc}
\usepackage[cmex10]{amsmath}
\usepackage{array}
\usepackage{subfig}
\usepackage{booktabs}
\usepackage{siunitx}
\usepackage{amsfonts}

\usepackage{algorithm}
\usepackage[noend]{algpseudocode}

\usepackage[hidelinks, bookmarks=false]{hyperref}

\usepackage{adjustbox}
\usepackage{multirow}
\usepackage{xcolor}
\usepackage{cite}
\usepackage{mathtools}

\usepackage[symbol]{footmisc}
\def\BibTeX{{\rm B\kern-.05em{\sc i\kern-.025em b}\kern-.08em
    T\kern-.1667em\lower.7ex\hbox{E}\kern-.125emX}}
\usepackage{graphicx}
\usepackage{pstricks}
\usepackage[utf8]{inputenc}
\usepackage{amssymb}
\usepackage{amsthm}
\usepackage{relsize}
\usepackage{bbm}
\usepackage{bm}
\usepackage[font={small}]{caption}
\usepackage{comment,color,soul}
\usepackage{textcomp}
\makeatletter
\def\thm@space@setup{\thm@preskip=2pt
        \thm@postskip=2pt \itshape}
\makeatother
\newtheoremstyle{newstyle}
{} 
{} 
{\mdseries} 
{} 
{\bfseries} 
{.} 
{ } 
{} 
\usepackage{url}
\theoremstyle{newstyle}
\theoremstyle{definition}

\theoremstyle{remark}

\IEEEoverridecommandlockouts

\renewcommand{\baselinestretch}{0.96}

\newcommand\rev[1]{\textcolor{black}{#1}}

\begin{document}

\title{Pre-defined Sparsity for Low-Complexity Convolutional Neural Networks}

\author{Souvik~Kundu,~\IEEEmembership{Member,~IEEE,}
        Mahdi~Nazemi,~\IEEEmembership{Member,~IEEE,}
        Massoud Pedram,~\IEEEmembership{Fellow,~IEEE,}\\
        Keith M. Chugg,~\IEEEmembership{Fellow,~IEEE,}
        and Peter A. Beerel~\IEEEmembership{Senior Member,~IEEE,}
\IEEEcompsocitemizethanks{\IEEEcompsocthanksitem S. Kundu, M.Nazemi, M. Pedram, 
                K. M. Chugg and P. A. Beerel are with Department of Electrical and 
                Computer Engineering, University of Southern California, Los Angeles, CA 90089, USA.
                E-mail: \{souvikku, mnazemi, pedram, chugg, pabeerel\}@usc.edu.
                
                \IEEEcompsocthanksitem This work was partly supported by NSF grant \#1763747 and AFRL award \#FA8750-16-2-0204 .
                
                \IEEEcompsocthanksitem Portions of this work were presented at the 57th Allerton Conference, 2019 \cite{kundu2019psconv}.
                }
        }

\IEEEtitleabstractindextext{
\begin{abstract} 
The high energy cost of processing deep convolutional neural networks impedes their ubiquitous deployment in energy-constrained platforms such as embedded systems and IoT devices. This work introduces convolutional layers with pre-defined sparse 2D kernels that have support sets that repeat periodically within and across filters. Due to the efficient storage of our periodic sparse kernels, the parameter savings can translate into considerable improvements in energy efficiency due to reduced DRAM accesses, thus promising significant improvements in the trade-off between energy consumption and accuracy for both training and inference.
To evaluate this approach, we performed experiments with two widely accepted datasets, CIFAR-10 and Tiny ImageNet in sparse variants of the ResNet18 and VGG16 architectures. Compared to baseline models, our proposed sparse variants require up to $\mathord{\sim}82\%$ fewer model parameters with $5.6\times$ fewer FLOPs with negligible loss in accuracy for ResNet18 on CIFAR-10.  For VGG16 trained on Tiny ImageNet, our approach requires $5.8 \times$ fewer FLOPs and up to $\mathord{\sim}83.3\%$ fewer model parameters with a drop in top-5 (top-1) accuracy of only $1.2\%$ ($\mathord{\sim}2.1\%$). We also compared the performance of our proposed architectures with that of ShuffleNet and MobileNetV2. Using similar hyperparameters and FLOPs, our ResNet18 variants  yield an average accuracy improvement of $\mathord{\sim}2.8\%$.
\end{abstract}

}

\maketitle
\IEEEdisplaynontitleabstractindextext

\section{Introduction}
\IEEEPARstart{I}n recent years, deep convolutional neural networks (CNNs) have become critical components in many real world vision applications ranging from object recognition \cite{krizhevsky2012imagenet, simonyan2014very, szegedy2015going, he2016deep} and detection \cite{girshick2014rich, sermanet2013overfeat, redmon2017yolo9000} to image segmentation \cite{tao2018image}. 
With the demand for high classification accuracy, current state-of-the-art CNNs have evolved to have hundreds of layers \cite{krizhevsky2012imagenet, simonyan2014very, coates2013deep, szegedy2015going, szegedy2016rethinking}, 
requiring millions of weights and billions of FLOPs.  
However, because a wide variety of neural network applications 
are heavily resource constrained, such as those for embedded and 
IoT devices, there is increasing interest in 
CNN architectures that balance implementation efficiency 
with accuracy and associated hardware accelerators 
that target CNNs \cite{han2016eie,chen2016eyeriss,chen2018eyeriss}. 
In particular, because energy is often the primary limited resource, 
researchers have focused on minimizing the number of 
non-zero model parameters and the accelerator's access 
to off-chip DRAM, which consumes around 200$\times$ 
more energy than access to on-chip SRAM \cite{chen2017using}.


Previous work has focused on accelerating inference and proposed {\em model pruning} \cite{yang2017designing, han2015learning, wen2016learning, liu2018rethinking, zhang2018systematic} and {\em quantization} \cite{zhou2017incremental, leng2018extremely, ren2019admm, alemdar2017ternary, courbariaux2015binaryconnect, rastegari2016xnor} to reduce the number of non-zero parameters. 
Recently, a more detailed analysis showed that such {\em unstructured pruning} may not reduce energy consumption because of the overhead required to manage sparse matrix representations \cite{wen2016learning}. 
\begin{figure*}[!t]
\includegraphics[width=0.8\linewidth]{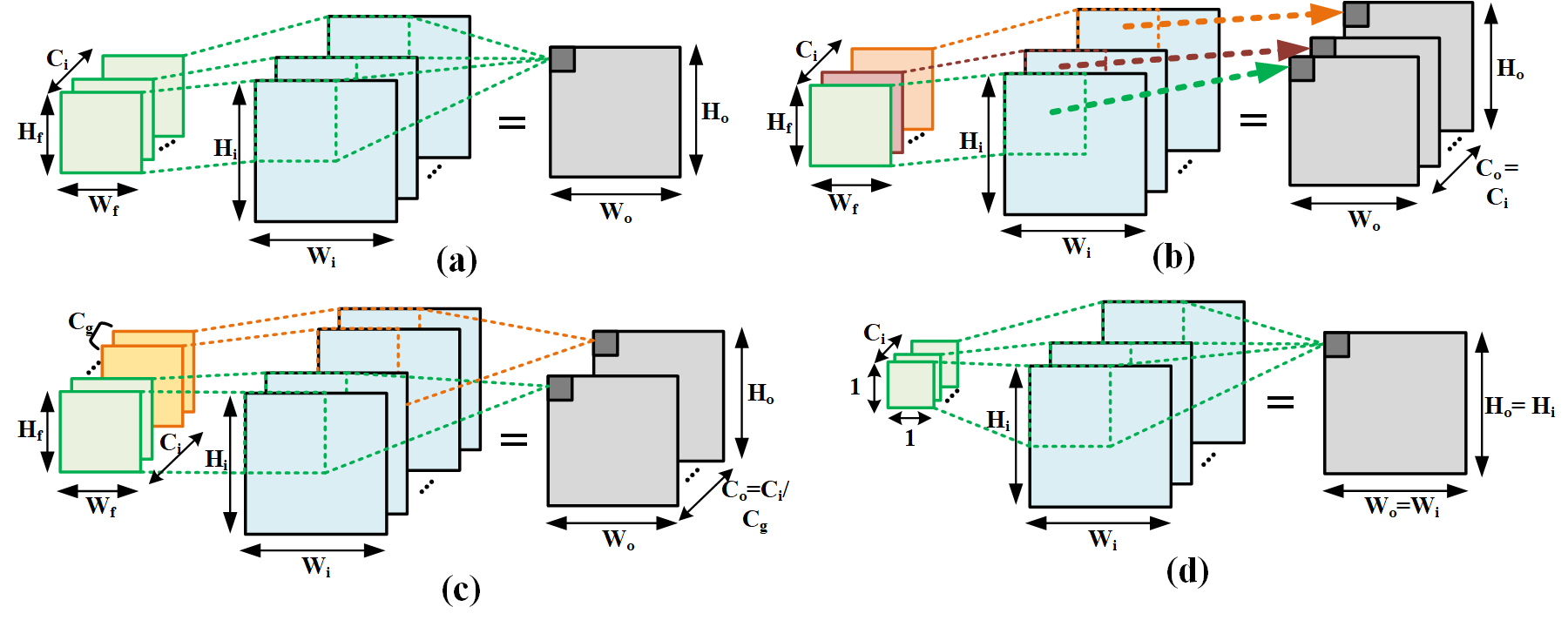}
\centering
\caption{Four major variants of convolutions: (a) standard fully connected convolution (SFCC), (b) depth-wise convolution (DWC), (c) group-wise convolution (GWC), and (d) point-wise convolution (PWC). }
\label{fig:diff_convs}
\end{figure*}
This motivates {\em structured pruning} \cite{wen2016learning}
which favors structure in the sparsity patterns that can more efficiently be managed in inference hardware.


Other work focused on the efficiency of both inference and training acceleration by defining notions of pre-defined sparsity \cite{dey2019pre, fayyazi2019csrram} in which a subset of the weights are fixed at zero before training and remain zero through inference.  
For example, a recent work \cite{dey2019pre} showed that neural networks can be trained with pre-defined hardware-friendly sparse connectivity in the fully connected multilayer perceptrons layers that avoids costly sparse matrix representations and thus can both speed-up and reduce the energy consumption of both inference and training. 
Other researchers have tried to address convolution (CONV) layers' computation complexity issue, which contribute the largest number of FLOPs for deep networks, exemplified by the CONV layer in ResNet18 \cite{he2016deep} which accounts for $\mathord{\sim}98\%$ of the total FLOPs for Tiny ImageNet classification. In particular many investigations have focused on efficient {\em pre-defined computationally-limited} 
filter designs to reduce complexity of training and inference
at the cost of accuracy, including MobileNet \cite{howard2017mobilenets}, MobileNetV2 \cite{sandler2018mobilenetv2}, and ShuffleNet \cite{zhang2018shufflenet}. 

This paper proposes pre-defined sparse convolutions
to improve energy and storage efficiency during both 
training and inference. We refer to this approach as 
{\em pSConv} and presented initial simulation results 
that show negligible performance degradation compared to 
fully-connected baseline models in \cite{kundu2019psconv}. 
However, as mentioned earlier, unstructured forms of pSConv may not lead to energy reductions due to the overhead of managing their sparse matrix representations.

Motivated by this fact, we extend pSConv by proposing a form of periodicity, repeating a relatively small pattern of
pre-defined sparse kernels within a 3D filter such that
fixed zero-weights occur repeatedly with a constant interval 
across the 3D filter. \rev{\textit{This periodicity can greatly reduce the overhead associated with managing sparsity, allowing the proposed CNN architecture to exhibit significant reductions in energy consumption compared to baseline CNNs with dense filters.}}

Finally, we present a convolutional channel modification to boost the accuracy of pSConv-based CNNs. In particular, the accuracy loss incurred due to the added periodicity constraint may be non-negligible  in some cases. To combat this phenomenon, we introduce fully-connected (FC) 2D kernels at fixed intervals within a 3D filter. \rev{\textit{In particular, extending the periodic pattern of pre-defined sparse kernels with a fully connected (FC) kernel boosts accuracy while maintaining relatively low storage overhead.}}

To evaluate the effectiveness of our proposed sparsity based CONVs, we run image classification tasks on variants of VGG \cite{simonyan2014very} and ResNet \cite{he2016deep} with CIFAR-10 \cite{krizhevsky2009learning} and Tiny ImageNet \cite{le2015tiny} datasets. 
We also show that we achieve higher test accuracy than MobileNetV2 \cite{sandler2018mobilenetv2} with similar network hyperparameter settings on these datasets. 
Finally, we analytically quantify the benefits of our algorithm compared to traditional approaches in terms of both FLOPs and storage, the latter assuming a variety of well-known sparse matrix representations. 

The remainder of this paper is structured as follows. Section \ref{sec:related_work} provides notable related work in the domain of CNN architectures and efficient sparse matrix representations. Section \ref{sec:our_work} describes our proposed architecture in detail and is followed by our analytical evaluation of FLOPs and storage requirements in Section \ref{sec:analytical}. We present 
our simulation results in Section \ref{sec:results} and conclude in Section \ref{sec:conc}. 

\section{Preliminaries and Related Work}
\label{sec:related_work}

CONV layers in neural network architectures transform the input images into abstract representations known as feature maps. To generate the output feature maps (OFMs) the filters of a layer are convolved with input feature maps (IFMs) which is comprised of the element wise product of filter and IFMs and the accumulation of partial sums. In particular, the following equation shows the computation of each OFM element in a standard fully-connected convolution (SFCC) layer. 

\begin{table}[b]
  \centering
  \caption{Descriptions of tensor dimensions in a convolutional layer}
  \begin{tabular}{|c|c|}
  \hline
    Variable &  Description \\\hline \hline
    $N$       & batch-size of a 3D feature map  \\\hline
    $H_i$, $W_i$ & height, width of IFM to a layer \\\hline
    $H_f$, $W_f $  & height, width of a 2D kernel in a layer \\ \hline
    $H_o$, $W_o$ & height, width of OFM from a layer \\\hline
    $C_i$ & \# of IFM channels/\# of 3D filter channels  \\ \hline
    $C_o$ & \# of OFM channels/\# of 3D filters \\\hline
    $C_g$ & \# of channels in a group from GWC 
    
    \\\hline
    $n$ & \# of parameters per kernel not pre-defined to be zero \\\hline
    \end{tabular}
  \label{tab:conv_vars}
\end{table}

\begin{align}\label{conv_op}
        O[z][v][x][y] = \mathrm{ReLU}\bigg(B[v] + \sum_{k=0}^{C_i -1}\sum_{i=0}^{H_f -1}\sum_{j=0}^{W_f -1} \nonumber\\  I[z][k][Sx+i][Sy+j]  
        W[v][k][i][j] \bigg) \nonumber \\
        0 \leq z < N, 0 \leq v < C_o, \nonumber \\ 0 \leq x < H_o, 0 \leq y < W_o
        \IEEEyesnumber
\end{align}
Here, {\em O}, {\em I}, {\em W} are the 4D OFM, IFM, and filter weight tensors, respectively and {\em B} is the 1D bias tensor added to each 3D filter result.
Also, $O[z][v][x][y]$ represents the $(x,y)^{th}$ OFM element in the $v^{th}$ output channel corresponding to the input batch $z$. 
Note the extensive data reuse both in IFM and weights, for which optimized dataflow is needed to ensure energy efficiency \cite{chakradhar2010dynamically, gupta2015deep, chen2017using}. 
The number of FLOPs  necessary to generate the OFM for a SFCC layer can be estimated as
\begin{align}\label{sfcc_flop}
    FL_{SFCC} = k^2  H_o  W_o C_o  C_i, \IEEEyesnumber
\end{align}
\rev{where, $k$ represents both height ($H_f$) and width ($W_f$) of the 2D kernel} and the meaning of the other variables are defined in Table \ref{tab:conv_vars}. \rev{Also, in this paper we assumed stride size of 1} and consider a FLOP and a multiply-accumulate operation to be equivalent.

\subsection{Pre-defined Computationally Limited Filters}

Because the SFCC \cite{lecun1999object}, shown in Fig. \ref{fig:diff_convs}(a), is computationally 
intensive, several pre-defined computationally-limited filters have been proposed to reduce the complexity of convolution.
These filters can be broadly classified into three different categories, as shown in Fig. \ref{fig:diff_convs}.
The first category is depth-wise convolution (DWC) \cite{vanhoucke2014learning}, shown in Fig. \ref{fig:diff_convs}(b). 
Here, each 2D kernel of size
$H_f \times W_f$ is convolved with a single channel of the IFM to produce the corresponding OFM; thus $C_i$ 2D kernels will produce an OFM of dimension $H_o \times W_o \times C_i$.  This requires $C_i$ times less computations compared to SFCC, but the output features capture no information {\em across} channels. 

The second category is group-wise convolution (GWC) \cite{krizhevsky2012imagenet}, shown in Fig. \ref{fig:diff_convs}(c),
which provides a compromise between SFCC and DWC. Here, 
a single channel of the OFM is computed by convolving
groups of $C_G$ channels from the IFM 
with partitions of the 3D filters, 
each of size $H_f \times W_f \times C_G$.   
Thus, with a total number of groups $G = C_i/C_G$, a 3D filter of dimension $H_f \times W_f \times C_i$ provides an OFM of size $H_o \times W_o \times G$.
Interestingly, SFCC can be viewed as GWC with $C_G = 1$ 
and DWC can be viewed as GWC with $C_G = C_i$.
Typically, the number of groups $G$ is chosen to be a small power of 2, but the choice is highly network architecture dependent \cite{ioannou2017deep}. 

Finally, Fig. \ref{fig:diff_convs}(d) illustrates PWC in which the 2D kernel dimension has size $1 \times 1$, thus generating a single OFM channel with low complexity. In particular, compared to a $3 \times 3$ 2D kernel dimension, the PWC has $9\times$ lower computational complexity. However, OFMs generated through this approach do not contain any embedded information {\em within} a channel.

Many well known network architectures have taken advantage of the benefits of pre-defined computationally-limited filters. For example, a combination of GWC and PWC was used in \cite{ioannou2017deep} and in the Inception modules \cite{szegedy2017inception, szegedy2016rethinking}. The ResNext  architecture \cite{xie2017aggregated} also uses a combination of GWC and PWC to replace each CONV layer of ResNet \cite{he2016deep}.  A class of scaled-down, reduced parameter architectures that replace most of the 3$\times$3 filters with PWC filters was dubbed SqueezeNet in \cite{iandola2016squeezenet}.
MobileNet and MobileNetV2, two popular variants of low complexity architectures designed to be implemented in mobile devices, replace the SFCC layer with a DWC followed by a PWC layer to gather information {\em across} channels. ShuffleNet \cite{zhang2018shufflenet} uses a combination of GWC, a channel shuffling for information sharing across 
channels, followed by a DWC layer.

\subsection{Sparse Matrix Storage Formats}

Most hardware platforms that process deep neural networks can benefit from sparse weight matrices only when such weights are represented through sparse matrix storage formats. 
These formats typically store non-zero elements of a given matrix in a vector while auxiliary vectors describe the locations of non-zero elements. 
This section explains three such methods commonly employed. 

\subsubsection{Coordinate List (COO)}

The COO format \cite{cheng2014professional} uses three vectors to represent a sparse matrix: a \textit{data} vector which keeps the values of non-zero elements, a \textit{row} vector which stores the row indices of non-zero elements, and finally, a \textit{column} vector which keeps track of column indices of non-zero elements.
For example, consider the sparse matrix $M$ shown below. 
\[
M = 
\begin{bmatrix}
0 & 1 & 0 & 0 & 2 & 0 & 3 \\
4 & 0 & 0 & 5 & 6 & 0 & 7 \\
0 & 0 & 0 & 8 & 9 & 0 & 0 \\
\end{bmatrix}
\]
The data, row, and column vectors for this matrix are as follows: 
\[
\mathrm{data} =
\begin{bmatrix}
1 & 2 & 3 & 4 & 5 & 6 & 7 & 8 & 9 \\
\end{bmatrix}
\]
\[
\mathrm{row} =
\begin{bmatrix}
0 & 0 & 0 & 1 & 1 & 1 & 1 & 2 & 2 \\
\end{bmatrix}
\]
\[
\mathrm{column} =
\begin{bmatrix}
1 & 4 & 6 & 0 & 3 & 4 & 6 & 3 & 4 \\
\end{bmatrix}
\]

In this representation, the size of all three vectors are the same and equal to the number of non-zero elements in the original sparse matrix.

\subsubsection{Compressed Sparse Row (CSR)}
Similar to the COO format, the CSR format \cite{cheng2014professional} uses three vectors to represent a sparse matrix. 
The \textit{data} vector stores values of non-zero elements in the order they are encountered when traversing the elements of the original matrix from left to right and top to bottom.
The \textit{column} vector keeps track of the column indices of non-zero elements, and the \textit{index} vector stores additional information used to identify the indices of the elements of each row of the matrix within the data vector. 
In fact, the column vector is the same as the one in the COO while the index vector stores the row vector in the COO format in a compressed manner, hence the name CSR. 
As an example, the data vector and the auxiliary vectors for the sparse matrix $M$ are as follows: 
\[
\mathrm{data} =
\begin{bmatrix}
\mathbf{1} & 2 & 3 & \mathbf{4} & 5 & 6 & 7 & \mathbf{8} & 9 \\
\end{bmatrix}
\]
\[
\mathrm{column} =
\begin{bmatrix}
1 & 4 & 6 & 0 & 3 & 4 & 6 & 3 & 4 \\
\end{bmatrix}
\]
\[
\mathrm{index} =
\begin{bmatrix}
0 & 3 & 7 & 9 \\
\end{bmatrix}
\]
Here, the bold entries in the data vector indicate the first nonzero elements of a new row of $M$ and occur at indices 0, 3, and 7 in the data vector. Thus, storing these indices in the index vector, along with the column vector, determine both the row and column for each element of the data vector.  The index vector always begins with zero and ends with the length of the data vector.   If a row of $M$ has no nonzero elements, the corresponding element in the index vector is repeated.  

\subsubsection{Compressed Sparse Column (CSC)}

The CSC format \cite{cheng2014professional} is very similar to the CSR and, in fact, is equivalent to CSR storage of the transpose of $M$. The column vector for CSR storage of $M^T$ is the \textit{row} vector for CSC storage of $M$ as follows
%
%
\[
\mathrm{data} =
\begin{bmatrix}
\mathbf{4} & \mathbf{1} & \mathbf{5} & 8 & \mathbf{2} & 6 & 9 & \mathbf{3} & 7 \\
\end{bmatrix}
\]
\[
\mathrm{row} =
\begin{bmatrix}
1 & 0 & 1 & 2 & 0 & 1 & 2 & 0 & 1 \\
\end{bmatrix}
\]
\[
\mathrm{index} =
\begin{bmatrix}
0 & 1 & 2 & 2 & 4 & 7 & 7 & 9 \\
\end{bmatrix}
\]
Similar to the CSR format, in the CSC format, the size of the data and row vectors are the same and equal to the number of non-zero elements. 
However, the size of the index vector is equal to the number of columns in the sparse matrix plus one. 

Some of the existing deep neural network (inference) accelerators such as Cambricon-X \cite{zhang2016cambricon}, \rev{Eyeriss \cite{chen2016eyeriss}\footnote{\rev{Note that Eyeriss \cite{chen2016eyeriss} uses run-length coding (RLC) to represent sparse vectors, in particular sparse activations, whereas $\mathrm{CSR}$ and $\mathrm{CSC}$ are better suited to represent sparse matrices.}}}, and Eyeriss~v2 \cite{chen2019eyeriss} have hardware support for processing values represented using sparse storage formats. 
The periodic sparsity introduced in this work allows us to further compress sparse representations such as the CSR and CSC formats by reusing the auxiliary vectors. 
This not only decreases the storage required for keeping model parameters in memory but also reduces the energy associated with transferring them from the main memory to processing elements (PEs). 
Furthermore, the proposed optimized sparse storage formats can be integrated into some of the existing accelerators such as Eyeriss~v2 with minor modifications to the controller logic or PEs. 
Section~\ref{sec:analytical} details the storage and energy savings achieved through deployment of the proposed formats. 
%

\section{Pre-defined Sparsity} 
\label{sec:our_work}

This section first describes pSConv, a form of pre-defined sparse kernel based convolution that we initially proposed in \cite{kundu2019psconv}.
It then describe how we introduce periodicity to this
framework to reduce the overhead of managing sparse matrix representations. Finally, the section presents a method to 
boost accuracy by periodically introducing a fully connected kernel 
into the 3D filters.

We define the {\em \textbf{kernel support}} as the set of 
entries in a $k\times k$ 2D kernel that are not constrained to be zero. The size of this set is
defined as {\em \textbf{kernel support size (KSS)}}.  The 
{\em \textbf{kernel variant size}} (KVS) is defined as the 
number of kernels with unique kernel support in a 3D filter.


\subsection{Pre-defined Sparse Kernels}
\begin{figure}[!ht]
\includegraphics[width=0.7\linewidth]{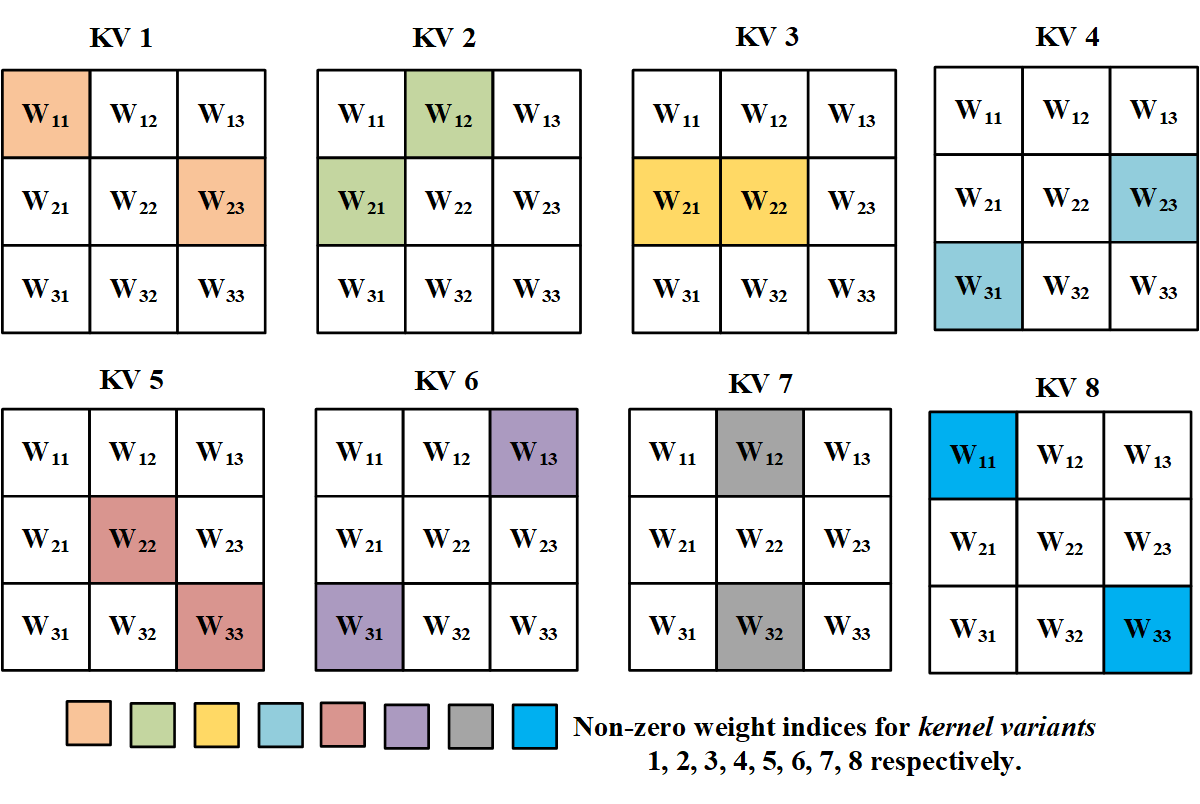}
\centering
\caption{An example of pre-defined sparse kernels with 8 different kernel variants each having KSS of 2.
The colored locations in each 2D kernel are allowed to have non-zero weight values.}
\label{fig:kernel_example}
\end{figure}
\begin{figure}[!t]
\includegraphics[width=0.8\linewidth]{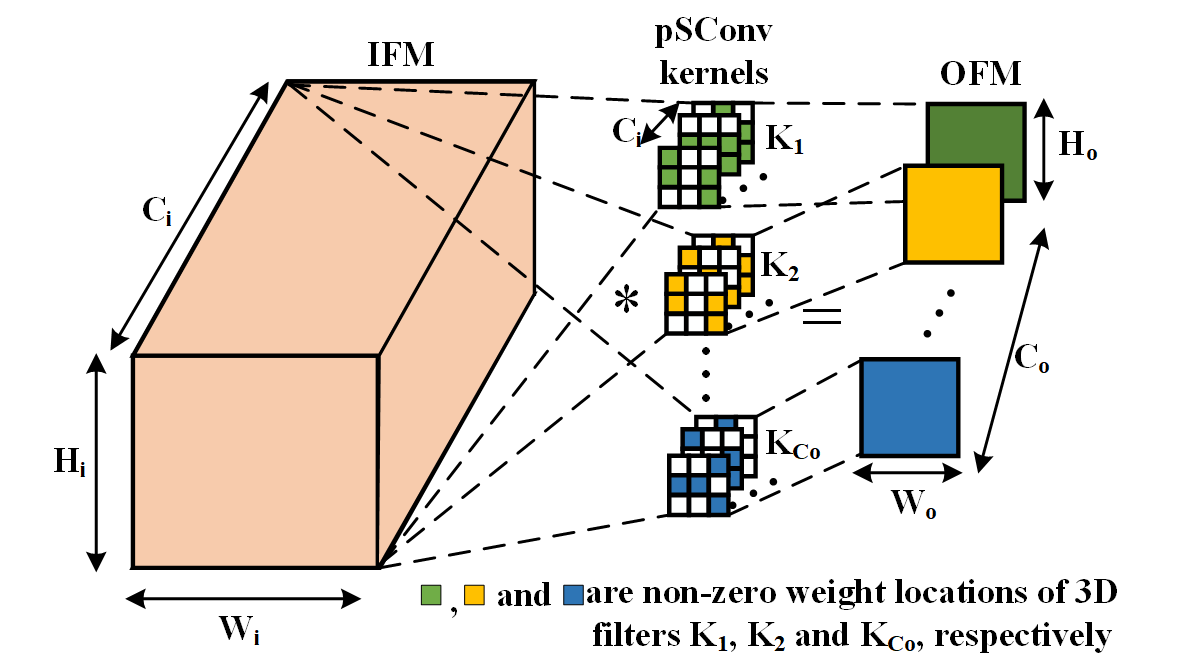}
\centering
\caption{An example of proposed  pre-defined  sparse  kernel  based  convolution with KSS of 4.}
\label{fig:pSConv_op}
\end{figure}

We say a 3D filter of size $k \times k \times C_i$ has pre-defined sparsity if some of the $k^2 \times C_i$ parameters are fixed to be zero before training and held fixed throughout training and inference. A {\em regular} pre-defined sparse 3D filter has the same KSS for each kernel that comprises the 3D filter.\footnote{We only consider the convolutional weights when defining sparsity. Bias and other variables associated with batchnorm are not considered because they add negligible complexity.} This regularity can help reduce the workload imbalance across different PEs performing multiply-accumulates and thus can help improve throughput of CNN accelerators \cite{chen2018eyeriss}. Fig. \ref{fig:kernel_example} shows an example of kernel variants. Here, $k = 3$, meaning $KSS = 9$ denotes the standard kernel without any pre-defined sparsity and $KSS = 2$ signifies that seven of the nine kernel entries are fixed at zero.
The choice of kernel variants can be viewed as a model search problem, however, in this paper we adopted a lower complexity approach of choosing them in a constrained pseudo-random manner which ensures every possible locations in $k^2$ 2D kernel space (9 in this case) has at-least one entry in a 3D filter which is not pre-defined to be zero. 
As an example, Fig. \ref{fig:pSConv_op} illustrates how an OFM of size $H_o \times W_o \times C_o$ is generated through convolution of $C_i \times C_o$ pre-defined sparse kernels of size $k^2$ with an IFM of size $H_i \times W_i \times C_i$.  

The challenge with efficiently implementing this scheme is how to avoid processing the weight entries that are fixed at zero. Because the {\em kernel variants} are chosen randomly from a potential set of $\binom{k^2}{KSS}$ options and KVS could be as large as $C_i$ for each 3D filter, the non-zero weight index memories can represent considerable overhead. 
We propose to address this problem by introducing periodicity within a 3D filter, as described below. 
 
\subsection{Periodic Sparse Kernel Patterns} 


\label{subsec:preodic_kernel}
 In order to reduce the overhead of storing the sparsity patterns, we propose to repeat the sparsity patterns, using only a small number of kernel variants across all filters.   
This is particularly beneficial in the compressed sparse weight formats because the same index values can be used for multiple filters.  
 
\begin{figure}[!ht]
\includegraphics[width=0.95\linewidth]{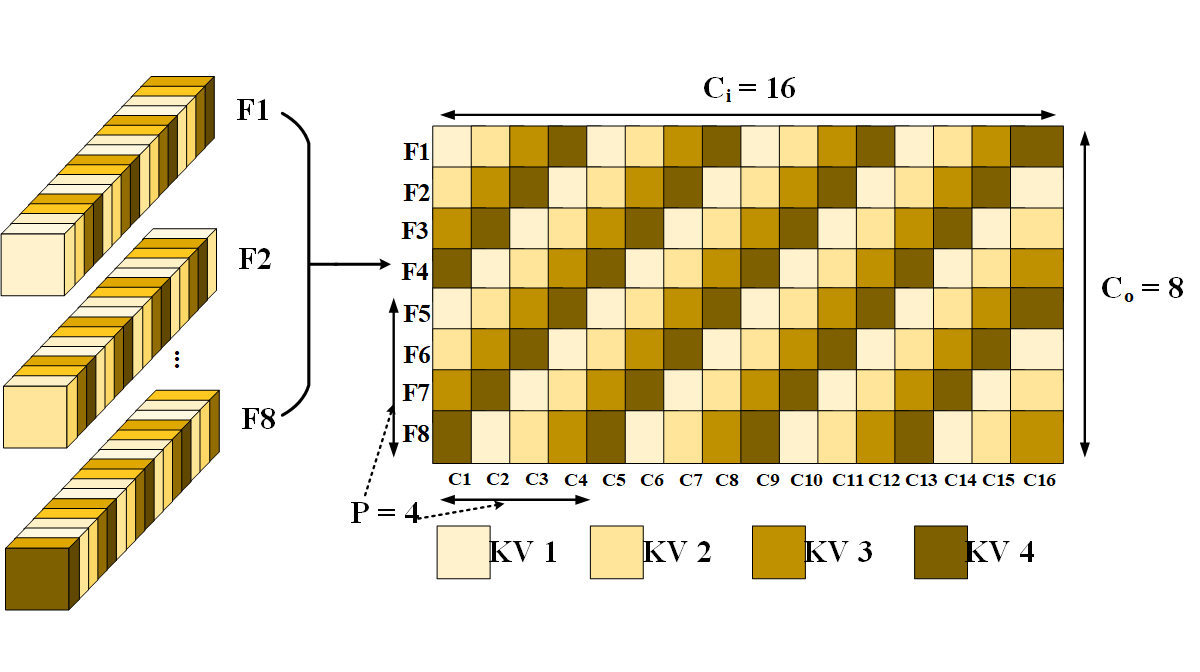}
\centering
\caption{Regular sparse kernel based 4D weight tensor. In the figure the 4D weight tensor has 4 different types of 2D kernel i.e. 4 different KVs (colored differently).}
\label{fig:periodic_sparse}
\end{figure}

 Fig. \ref{fig:periodic_sparse} shows an example of periodically repeating kernel patterns, with a periodicity $P = KVS = 4$. 
Notice to retain periodicity across different 3D filters and while still providing some diversity, we rotate the sequence of kernel variants, starting each filter (of $P$ consecutive filters) with a different kernel variant. 
For instance, if the first 3D filter starts with KV1 followed by KV2, KV3, and KV4, and then repeats the order, we start the second 3D filter with KV2 to create a repeating sequence of [KV2, KV3, KV4, KV1]. Thus, we maintain the sequence of repeating kernels modulo rotation. 

\rev{Our specific choice of sparse KVs in our experiments are obtained by sequentially picking non-zero 2D entries randomly constrained such that no non-zero 2D entry is chosen twice until all entries of the kernel are chosen at least once.  Furthermore, we ensure that every pixel in the input frame has an opportunity to affect the outcome of our sparse-periodic network which
constrains the minimum value of periodicity $P$. For example, for a $3\times 3$ kernel, with $KSS$ of 1, the minimum value of $P$ necessary to ensure every entry of the kernel is chosen is 9. More specifically, the nine sparse $3 \times 3$ 2D kernels in this example each must have a different single non-zero entry such that together they cover all entries.}

\subsection{Boosting Accuracy with FC Kernels}

\begin{figure}[!ht]
\includegraphics[width=0.8\linewidth]{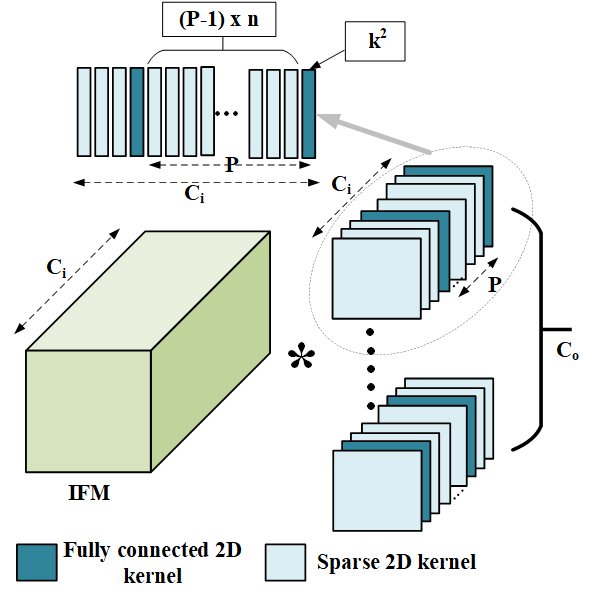}
\centering
\caption{Periodic insertion of FC 2D kernels between
sparse kernels.}
\label{fig:periodic_dense}
\end{figure}

Although the periodicity in sparse patterns is beneficial for  overhead management of the sparsity, the choice of KSS and the simplistic way of choosing {\em kernel variants} may sometimes cost significant classification performance. Methods to find suitable sparse patterns and KVS values through pattern pruning, inspired by image smoothing filters, were recently considered in  \cite{ma2019pconv}.  However, here, we propose a complementary approach in which we periodically introduce fully connected (FC) kernels, i.e., kernels with KSS = $k^2$, within each 3D filter. \rev{We use $\eta$ to denote the number of dense or FC kernels in a period $P$ and, in principle, it can have any value between 0 and $P$. In the case of standard (dense) convolution filter based models, $\eta = P$. In contrast, $\eta = 0$ implies no boosting.} 
Fig. \ref{fig:periodic_dense} illustrates the idea where one fully connected kernel ($\eta$ = 1) is introduced every $P$ kernels and
with other $P-1$ sparse kernels, repeating
this $P$ 
pattern throughout the 3D filter. 

\rev{Note that our selection of the period $P$ is premised on the fact that balancing the computational requirements across 3D filters is preferred for hardware implementations because it enables more efficient scheduling across parallel computational units. It is therefore desirable to have a fixed number of non-zero weights per filter which implies having an equal number of FC kernels per filter. Given the approach illustrated in Fig. \ref{fig:periodic_sparse} it is therefore preferred to have $C_i$ be divisible by $P$. In our experiments, detailed in Section \ref{sec:results}, the layers where boosting is applied have $C_i$ $\in \{ 64, 128, 256, 512\}$. Thus our preferred values of $P$ are \{2, 4, 8, 16, 32, 64\}.}

\rev{To choose the sparse kernel variants we follow the same principle as described in Section \ref{subsec:preodic_kernel} before adding the $\eta$ FC kernels. However, in the presence of an FC kernel, $P$ can, in principle, be lower than the minimum $P$ without boosting because the FC kernel covers all entries.}  
\rev{Moreover, when} $KVS$ $<$ $P$, we propose to randomly reuse some of the sparse kernel variants to maintain periodicity.

\section{FLOPs and Energy Efficiency Analysis}
\label{sec:analytical}

\subsection{Complexity Analysis}
\label{subsec:flops}

\begin{table}[!ht]
  \centering
  \caption{Expression of FLOPs count for inference operation with various pre-defined computationally-limited filters}
  \begin{tabular}{|c|c|}
  \hline
  Approach & FLOP count (forward, ideal) \\\hline \hline
  MobileNet-like \cite{howard2017mobilenets} & $H_o   W_o   C_i (k^2 + C_o)$ \\ 
  (DWC+PWC) &  \\\hline
  ShuffleNet-like \cite{zhang2018shufflenet} & $H_o  W_o   C_o  C_i(\frac{k^2}{G}  + 1)$ \\ 
  (GWC+PWC) & \\\hline
  \end{tabular}
  \label{tab:flops}
\end{table}

    The total FLOPs for MobileNet-like and ShuffleNet-like CONV layers can be estimated as shown in Table \ref{tab:flops}. 
    The total FLOPs for sparse (both periodic and aperiodic variants) kernel based CONV layers with KSS of $n$ can be estimated as
    
    \begin{align}\label{flops3}
    FL_{S} = H_o  W_o  C_i C_o n. \IEEEyesnumber
    \end{align}
    To estimate the FLOPs of sparse kernel based CONVs with boosting\footnote{\rev{Here each period is assumed to have only one FC kernel}.}, we start with the number of elements in a period ($P$) that are allowed to be non zero
    which can be computed as (shown in Fig. \ref{fig:periodic_dense}),
    \begin{align}\label{flops1}
        W_{P} = (P-1) n + k^2 \IEEEyesnumber
    \end{align}
    Now, with total number of $(C_i/P)$ dense, and $(C_i -C_i/P)$ sparse kernels in each 3D filter of a layer the FLOPs can be computed as,
    \begin{align}\label{flops4}
       FL_{\mathrm{PSD}} =  \left[\left({\frac{C_i}{P}}\right) k^2 + \left(C_i - \left({\frac{C_i}{P}}\right)\right)n\right] H_o W_o  C_o \IEEEyesnumber
    \end{align}
    
    The ratio of the FLOP counts for MobileNet-like and ShuffleNet-like layers to that of sparse kernel based CONVs with boosting is 
    \begin{align}
        R_{\mathrm{mob}} &=  \frac{ \text{FLOPs for MobileNet-like}}{ \text{FLOPS for periodic-sparse with boosting}} \nonumber  \\
        &= \frac{P(k^2 + C_o)}{[k^2 + (P-1)  n]   C_o}
        \IEEEyesnumber
        \label{eq:ratio_mob}
    \end{align}
    
    \begin{align}
        R_{\mathrm{shuf}} &= \frac{ \text{FLOPs for ShuffleNet-like}}{ \text{FLOPS for periodic-sparse with boosting}}\nonumber \\
        &= \frac{ P (\frac{k^2}{G} + 1)}{[k^2 + (P - 1)  n] }
        \IEEEyesnumber
        \label{eq:ratio_shuff}
    \end{align}
    
\begin{figure}[h!]
\includegraphics[width=0.99\linewidth]{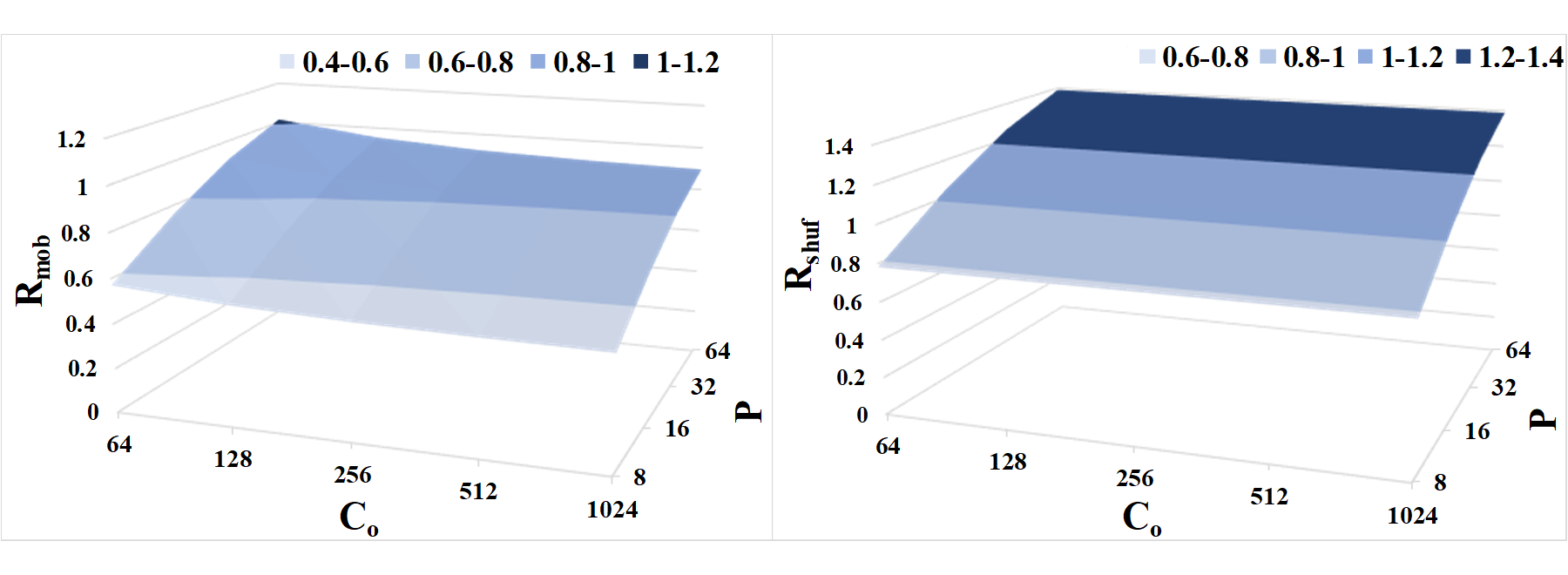}
\centering
\caption{\rev{A 3D illustration of the change in $R_{\mathrm{mob}}$ and $R_{\mathrm{shuf}}$ as a function of the $C_o$ and $P$. Here we assumed $G$, $k$, and $n$ to be 16, 3, and 1, respectively}.}
\label{fig:3D_R_mob_shuf}
\end{figure}

    It is clear that we will have computational saving when the values of  
    $R_{\mathrm{mob}}$ and $R_{\mathrm{shuf}}$ are greater than 1. When $C_o$ is large and $P >> \frac{k^2}{n}$, 
     (\ref{eq:ratio_mob}) and (\ref{eq:ratio_shuff}) can be approximated as
    \begin{align}
        & R_{\mathrm{mob}} \simeq \frac{1}{n} \\\IEEEyessubnumber 
        & R_{\mathrm{shuf}} \simeq \frac{(\frac{k^2}{G} + 1)}{n} \IEEEyessubnumber
    \end{align}
    which shows the complexity increment due to periodic insertion of FC kernels is negligible for relatively wide networks with large periods. \rev{Fig. \ref{fig:3D_R_mob_shuf} shows a 3D illustration of the per layer FLOP ratios ($R_{\mathrm{mob}}$ and $R_{\mathrm{shuf}}$) as a function of $C_o$ and $P$. Note that even though the per layer ratio can be less than 1, the total parameter count for MobileNet or ShuffleNet-like networks can be larger due to the presence of more layers}.
    
\subsection{The Impact of Periodicity on Storage and Energy}
\label{sec:evalofmerit}
Sparsity leads to savings in storage only when the overhead of storing the auxiliary vectors to manage sparsity is negligible. 
This section presents a new sparse representation specifically tailored to periodic sparse kernels and compares it with existing formats. 
It also analyzes storage requirements of different sparse representations analytically, allowing the study of the effectiveness of such formats at different levels of density. 
Furthermore, it explains how the proposed representation can be exploited in CNN accelerators. 

\subsubsection{CSR/CSC with a Periodic Column/Row Vector} \label{subsec:periodic_smsf}

The periodic pattern of kernels introduced in Section~\ref{subsec:preodic_kernel} allows reusing the column/row vector in the CSR/CSC format. 
For example, assume a convolutional layer with $3 \times 3$ kernels, 128 input channels, 128 output channels, and a period of four. 
The 4D weight tensor corresponding to this convolutional layer can be represented by a \textit{flattened weight matrix} where each row corresponds to a flattened filter. 
As a result, the number of rows in the flattened weight matrix is equal to 128 while the number of columns is $3 \times 3 \times 128 = 1152$. 
Because of the periodicity across filters, the structure of the rows of the flattened weight matrix will also repeat with a period of four. 
Therefore, one can simply store the column vector of the CSR format for the first four rows and reuse them for the subsequent rows. 
We refer to this new sparse storage format as CSR with a periodic column vector and denote it with $\mathrm{CSR_P}$, where $P$ denotes the period of repetition of the column vector. 

Similarly, because of the periodicity of kernels within a filter, the columns of the flattened matrix  also repeat with a period of $4 \times (3 \times 3) = 36$. 
As a result, one can choose to use the CSC format to represent the flattened sparse matrix and reuse the row vector for groups of 36 columns. 
We refer to this new format as CSC with a periodic row vector and denote it with $\mathrm{CSC_{P}}$, where the $P$ here denotes the period of repetition of the row vector. 

Table~\ref{tab:smsf_vars} summarizes the notation used for comparing the storage cost of different storage formats. 
Using the notation introduced here, Table~\ref{tab:smsf_comparison} explains storage requirements of different storage formats. 
\begin{table}[ht]
  \centering
  \caption{Summary of notation for matrix storage formats}
  \begin{tabular}{|c|c|}
  \hline
    Variable &  Description \\\hline \hline
    $H_F$, $W_F$ & height, width of a flattened weight matrix \\\hline
    $\rho$ & density ($0 \leq \rho \leq 1$) \\\hline
    $b_v$ & number of bits for representing data values \\\hline
    $b_r, b_c$ & number of bits for representing row, column values \\\hline
    $b_i$ & number of bits for representing index values \\\hline
    $b_P$ & number of bits for representing the period \\\hline
    \end{tabular}
  \label{tab:smsf_vars}
\end{table}
\begin{table}[ht]
  \centering
  \caption{Storage requirement of storing a matrix using dense and sparse storage formats}
  \begin{tabular}{|c|c|}
  \hline
    Format &  Storage Requirement (bits) \\\hline \hline
    Dense & $H_F W_F b_v$ \\\hline
    COO & $\rho H_F W_F (b_v + b_r + b_c)$ \\\hline
    CSR & $\rho H_F W_F (b_v + b_c) + (H_F + 1) b_i $ \\\hline
    CSC & $\rho H_F W_F (b_v + b_r) + (W_F + 1) b_i $ \\\hline
    $\mathrm{CSR_P}$ & $\rho H_F W_F b_v + \rho P W_F b_c + (H_F + 1) b_i + b_P $ \\\hline
    $\mathrm{CSC_{P}}$ & $\rho H_F W_F b_v + \rho P H_F b_r + (W_F + 1) b_i + b_P $ \\\hline
    \end{tabular}
  \label{tab:smsf_comparison}
\end{table}

Based on Table~\ref{tab:smsf_comparison}, the COO format is expected to have higher overhead than that of the CSR and CSC formats, which have similar storage overhead. 
Furthermore, it is evident that the introduction of periodicity to the CSR and CSC formats can significantly decrease the storage overhead. 
%
%

\subsubsection{\rev{Application to Weight Sub-Matrices}}

\begin{figure}[tb]
    \centering
    \includegraphics[width=1\columnwidth]{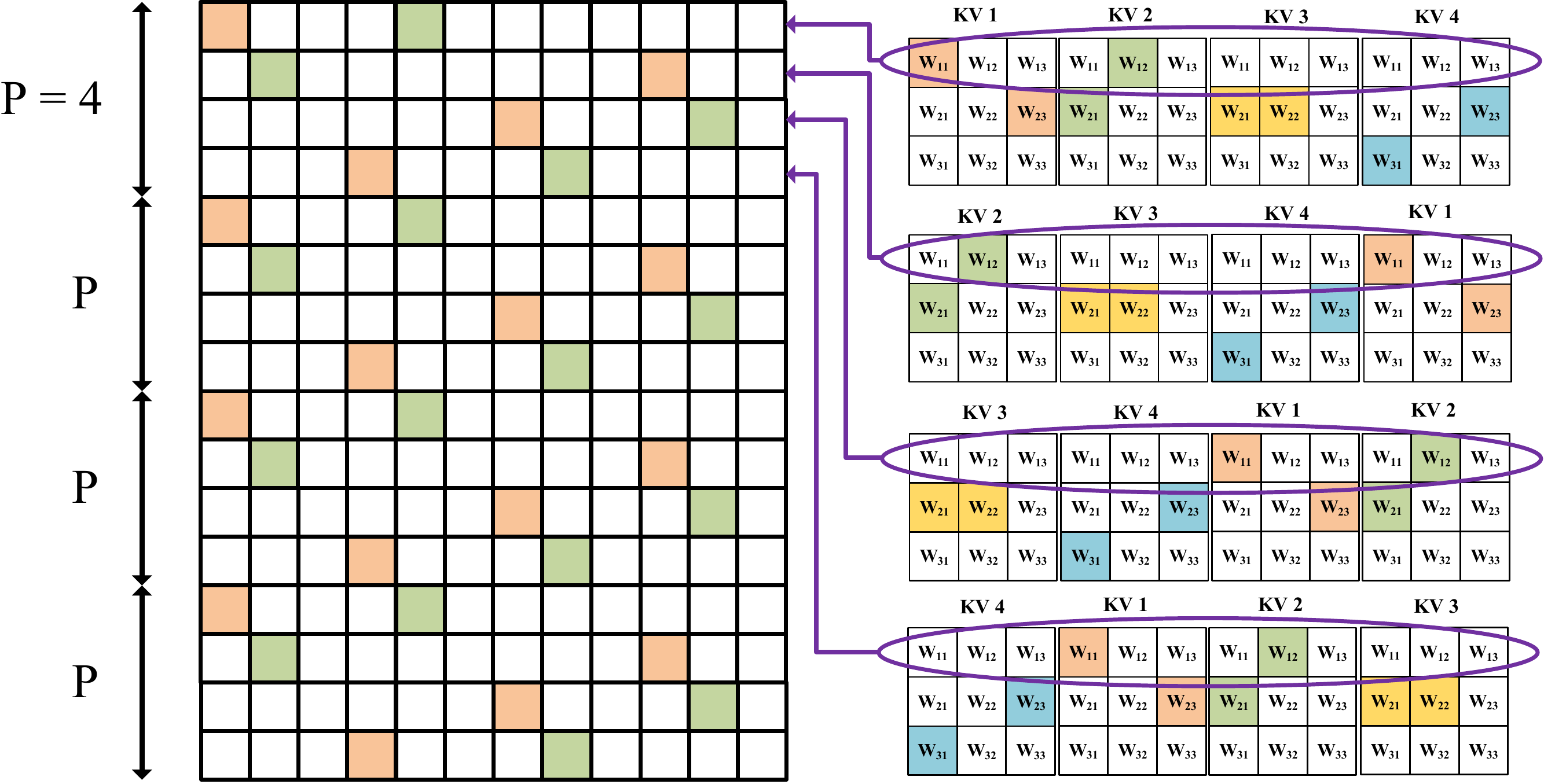}
    \caption{Illustration of how periodicity in a filter leads to repeating rows of sub-matrices of the filter's flattened weight matrix. 
    }
    \label{fig:periodic_sparse_csr}
\end{figure}

%
As noted above, a convolutional layer with periodic sparse kernels induces a flattened weight matrix that also has periodically repeating columns and rows.  
In a CNN accelerator, the processing of a convolutional layer is often broken down into smaller operations where subsets of the flattened weight matrix are processed across multiple PEs.  This processing requires accessing a sub-matrix of the flattened weight matrix.  If this sub-matrix is large enough, it will also have row or column vectors that are repeated periodically.   
For example, Fig.~\ref{fig:periodic_sparse_csr} demonstrates a subset of a flattened weight matrix that is used in a single processing element of an architecture like Eyeriss~v2 \cite{chen2018eyeriss} (the original flattened weight matrix is built using the first four kernel variants shown in Fig.~\ref{fig:kernel_example}). 
This sub-matrix corresponds to processing the first (top) row of four kernels of 16 filters.  
Specifically, the  sub-matrix consists of 16 rows corresponding to 16 filters and 12 columns corresponding to the top row of four kernels per filter.  Note in Fig.~\ref{fig:periodic_sparse_csr}, the four kernels have been rotated as described in Section~\ref{subsec:preodic_kernel}.  
Based on the  periodic pattern across filters, the sub-matrix shown in Fig.~\ref{fig:periodic_sparse_csr} has repeating rows with a period of four and can be represented using $\mathrm{CSR_4}$. 
%

%
%

\begin{figure*}[tb]
    \centering
    \subfloat[]{
        \includegraphics[width=0.45\textwidth]{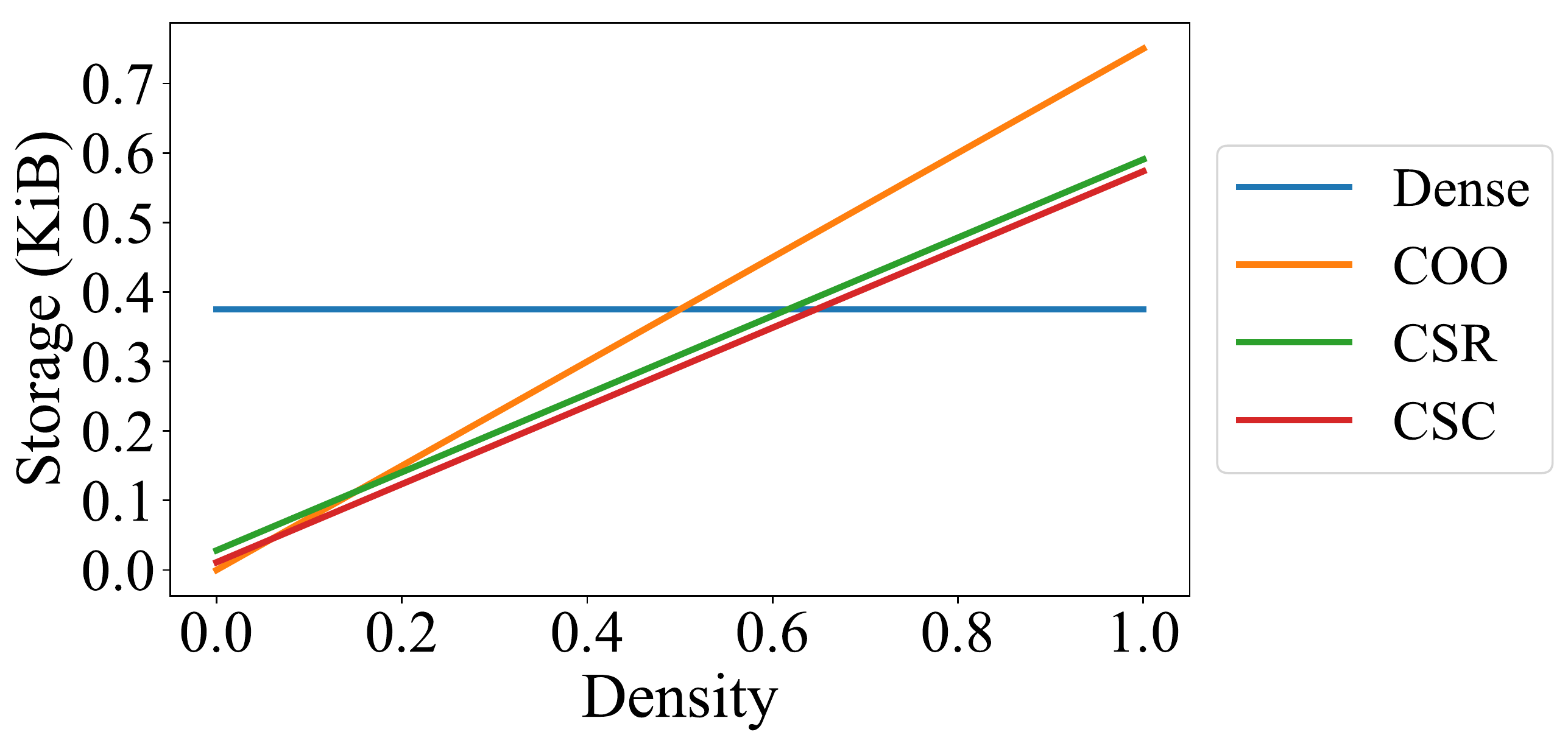}
        \label{fig:eyeriss_v2_sparse}
    }
    \subfloat[]{
        \includegraphics[width=0.45\textwidth]{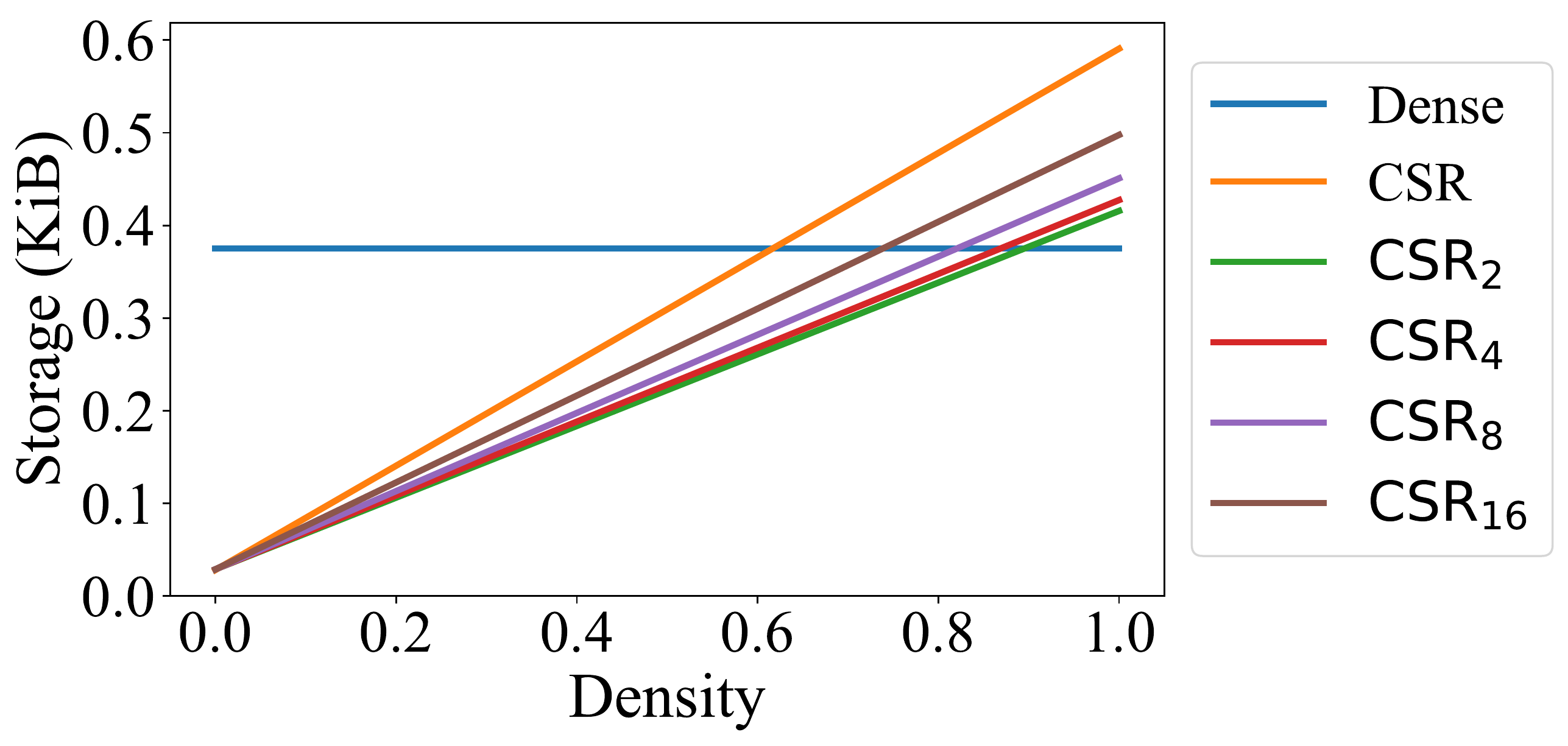}
        \label{fig:eyeriss_v2_sparse_periodic}
    }
    
    \caption{Comparison of storage requirements of (a) various existing storage formats and (b) dense, CSR, and $\mathrm{CSR_P}$ formats at different levels of density for a matrix of size $32 \times 12$ ($b_v = 8$, $b_c = b_r = 4$, $b_i = 7$, and $b_P=6$).
    }
\end{figure*}

\rev{Because each PE in a CNN accelerator processes a small portion of the flattened weight matrix, $b_c$, $b_r$, and $b_i$ have small ranges and therefore can be represented using a small number of bits. }
For example, assuming $b_v = 8$, $b_c = b_r = 4$, and $b_i = 7$, Fig.~\ref{fig:eyeriss_v2_sparse} compares storage requirements of various existing storage formats at different levels of filter density. 
It is observed that the CSR and CSC formats yield lower total storage when the original matrix is at most 62\% and 65\% dense, respectively. 
Fig.~\ref{fig:eyeriss_v2_sparse_periodic} compares storage requirement of dense, CSR, and $\mathrm{CSR_P}$ formats for the same matrix that was shown in Fig.~\ref{fig:eyeriss_v2_sparse}, for different values of $P$, and $b_P = 6$. 
It is observed that the $\mathrm{CSR_8}$ and $\mathrm{CSR_{16}}$ yield lower total storage when the original matrix is at most 82\% and 73\% dense, respectively. 
Furthermore, at 62\% density, $\mathrm{CSR_8}$ and $\mathrm{CSR_{16}}$ yield lower total storage compared to $\mathrm{CSR}$ by 23\% and 16\%, respectively\rev{\footnote{\rev{Interestingly, $\mathrm{CSR}$ has similar storage requirements as  $\mathrm{RLC}$. In particular, as implemented in Eyeriss \cite{chen2016eyeriss}, at 62\% density, $\mathrm{RLC}$ would lead to 0.14\% more storage than $\mathrm{CSR}$.}}}. 
This is equivalent to 60.04\% and 39.86\% reduction in the overhead of storing auxiliary vectors for the $\mathrm{CSR_8}$ and $\mathrm{CSR_{16}}$ compared to the $\mathrm{CSR}$ format, respectively. 

\rev{Because the energy cost associated with transferring from the DRAMs is well-modeled as proportional to the number of bits read \cite{greenberg2013lpddr3}, the reduced storage requirements of $\mathrm{CSR_P}$/$\mathrm{CSC_{P}}$ lead to a proportional reduction in the energy cost associated with DRAM access.  For example, a 50\% savings in storage will result in a $\mathord{\sim}2\times$ reduction in energy consumption related to DRAM access. For this reason, in the remainder of this paper, we focus on savings in storage requirements with the energy savings being implicit. }  

\subsubsection{Hardware Support for Periodic Sparsity}

%
%

\rev{The low-complexity storage formats introduced in Section~\ref{subsec:periodic_smsf}, i.e.  $\mathrm{CSR_P}$/$\mathrm{CSC_{P}}$, cannot be integrated into existing accelerators without ensuring they can support the proposed periodic sparse format. }
For example, in Eyeriss~v2, each weight value (i.e. data) is coupled with its corresponding index and they are read as a whole from the main memory. 
On the other hand, the $\mathrm{CSR_P}$/$\mathrm{CSC_{P}}$ store the column/row vector separately from the data vector and read the auxiliary vectors once for all data values. 
This not only requires proper adjustment of the bus that transfers data from the DRAM to the chip but also may require a minor modification in either the control logic or PEs. 

One approach to make an accelerator like Eyeriss~v2 compatible with periodic sparsity is to store the weights in DRAM using the proposed sparse periodic format and modify the system-level control logic to expand the column/row vector before storing them in the PE's scratchpad memory. 
In other words, the sparse column/row vector is read from the DRAM only once, but replicated before being written into the scratchpad memory corresponding the the column/row vector so that they adhere 
to the CSR/CSC format. 
In this manner, the scratchpad memory within each PE remains the same and stores bundled (data, index) pairs. Because DRAM accesses consume two orders of magnitude more energy than on-chip communication, we can thus achieve close to
the optimal energy savings without requiring any 
change in the PE array or its associated control 
structures.

A more comprehensive approach to supporting periodic sparsity involves ensuring the PEs can use the column/row scratchpad memory as a configurable circular buffer, which, to support periodicity, will be configured to have length $P$. 
This type of support may already exist because in many cases, the size of the weight matrix processed within each PE is smaller than the size of the corresponding scratchpad memory and therefore, only a portion of the scratchpad memory is used. 
In this approach, the periodic column/row vector is read from the DRAM once, written into the scratchpad memory, and accessed multiple times for different rows of the weight matrix. This reduces the required on-chip communication and thus may save more memory 
compared to storing the expanded column/row vectors 
in the scratchpad memory.

While the presented approaches enable compression of the column/row vectors, one may be able to compress the index vector as well, as suggested by the row periodicity illustrated in Fig.~\ref{fig:periodic_sparse_csr}.  
However, this may require more complex hardware support to expand the index vector before storing them in the PEs or adding support for the compressed index vectors within the PE. 

\section{Experimental Results}
\label{sec:results}

\begin{table}[h]
 \centering
 \caption{Nomenclature of the network architectures used in simulation}
  \begin{minipage}{\columnwidth}
   \resizebox{\columnwidth}{!}{
   \begin{tabular}{|c|c|}
    \hline
    Name & Description of the network architecture \\\hline
    aaa\_pSC<n> & aaa network with pre-defined sparse  \\                      
                & kernel based convolution where each 2D kernel\\
                &  has n weights not pre-defined to be zero.\\ \hline
    aaa\_pSC<n>\_P<m> & aaa network with every $m^{th}$ kernel is \\
                    & FC and rest are pre-defined sparse kernels  \\     
                    & having n weights not pre-defined to be zero.\\\hline
    aaa\_PS<n>\_P<m> & aaa network with both periodicity \\
                 & and kernel variant values of m, and each 2D kernel  \\
                 & has n weights not pre-defined to be zero. \\ \hline
    aaa\_PSD<n>\_P<m> & aaa network with periodic kernel variants\\
                        & having periodicity m, where each period has $m-1$\\
                        & sparse kernel variants each with n weights not pre-\\
                        & defined to be zero and 1 FC $k \times k $ kernel. \\\hline
    \end{tabular}
    }
   \end{minipage}
 \label{tab:nomenclature}
\end{table}
This section describes our simulation results and analysis.
We first detail the datasets, architecture, and important hyperparameters used for our experiments, followed by our experimental results of our proposed pSConv approach, the introduction of periodicity, and our performance boosting technique. Finally, we compare our modified network architectures with MobileNetV2 \cite{sandler2018mobilenetv2}, a popular low-complexity CNN variant for image classification, in terms of FLOPs, model parameters, and accuracy.  
We used Pytorch \cite{paszke2017automatic} to design the models and trained/tested the models on AWS EC2 P3.2x large instances that have an NVIDIA Tesla V100 GPU. 

\subsection{Datasets, Architectures, and Hyperparameters}
\label{subsec:data_arch_hyper}

To evaluate our models we used CIFAR-10 \cite{krizhevsky2009learning} and Tiny ImageNet \cite{le2015tiny}, two widely popular image classification datasets. The input image dimensions of CIFAR-10 and Tiny ImageNet are ($32 \times 32 \times 3$) and ($ 64 \times 64 \times 3$), respectively. The number of different output classes for these two datasets are 10 and 200, respectively.  
We chose variants of VGG16 \cite{simonyan2014very} and ResNet18
\cite{he2016deep} as the base network models to apply our architectural modifications. The VGG16 architecture has thirteen $3 \times 3$ kernel based convolutional layers. The flattened output of final CONV layer is fed to the fully connected part having three fully connected (FC) layers.\footnote{In VGG16 for CIFAR-10 dataset, we used only one FC layer because the input image dimension is $4 \times$ smaller than Tiny ImageNet and multiple FC layers are not needed to achieve high accuracy.} The CONVs of ResNet18 architecture
consists of four layers each containing two basic blocks, where each basic block 
has two convolutional layers along with a skip connection path.    
We used pre-defined sparse kernels on all $k \times k$ CONV layers where $k > 1$ but excluded the first layer, as it is connected to the primary inputs and is thus more sensitive to zero weights. 
Training was performed for 120 and 100 epochs for CIFAR-10 and Tiny ImageNet, respectively. The initial learning rate was set to 0.1 with momentum of 0.9 and weight decay value to $5 \times 10^{-4}$. The image datasets were augmented through random cropping and horizontal flips before being fed into the network in batches of 128 and 100 for CIFAR-10 and Tiny ImageNet, respectively. All results reported are the average over two training experiments. Table \ref{tab:nomenclature} provides the names of each variant of network model and corresponding architecture descriptions.
%

\begin{figure*}[h!]
\includegraphics[width=0.8\linewidth]{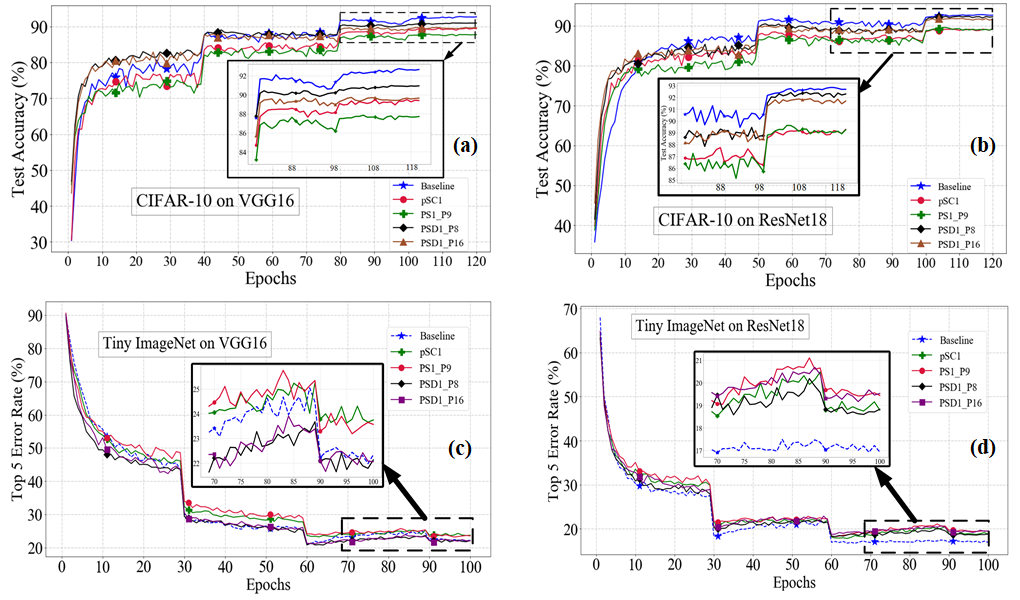}
\centering
\caption{(a), and (b) shows the test accuracy vs. epochs for CIFAR-10 dataset in different variants of VGG16 and ResNet18 models, respectively; (c), and (d) are plots of top 5 error rate vs. epochs for Tiny ImageNet dataset in different variants of VGG16 and ResNet18 models, respectively. The KSS for all the variants is 1. }
\label{fig:Acc_error_cifar_tiny_res_vgg}
\end{figure*}

\subsection{Results for pSConv Based CNN}
\label{subsec:psconv}

We analyzed three different variants of regular sparse kernel based CONVs with KSS values of 4, 2 and 1 along side  the baseline standard convolution based network. As stated earlier, in our choice of kernel patterns we ensure each of the $k^2$ possible kernel entries are covered by at least one
sparse kernel variant. Table \ref{tab:vgg_res_pSC_cifar_tiny} provides the results in terms of accuracy and parameter count\footnote{We considered the convolution layer parameters only to report in the tables of this section without considering the overhead of indexing.} with the KSS variants applied in VGG16 and ResNet18 architectures. The ResNet18-based results show that even with KSS of only 4, the test accuracy degradation is within $\mathord{\sim}0.4\%$ for CIFAR-10 dataset, and within $\mathord{\sim}0.6\%$ for Tiny ImageNet. The same results for VGG16 show a test accuracy degradation is within $\mathord{\sim}0.7\%$ for CIFAR-10 dataset, and within $\mathord{\sim}1.1\%$ for Tiny ImageNet.
\begin{table}[h!]
  \centering
   \caption{Test accuracy of pSConv based  VGG16, and ResNet18 on CIFAR-10 and Tiny ImageNet. Here we use KSS of 9, 4, 2, and 1, respectively. Also, KSS of 9 means SFCC based CONVs and thus they are used as baseline to compare accuracy, and parameters.}
   \begin{minipage}{\columnwidth}
    \resizebox{\columnwidth}{!}{
    \begin{tabular}{|c|c|c|c|c|c|}
    \hline
    Data & Model &      Top 1   & Top 5             &  Parameters & Parameters  (\%) \\
    set &       &   acc $(\%)$  &     acc $(\%)$    &             &   reduction \\ \hline
        &  VGG16$\_$pSC9   & \textbf{92.8} & --  & 14.73 M & --- \\
    \cline{2-6}    
     C  &  VGG16$\_$pSC4   & 92.0 & -- &  6.55 M & 55.56 \\
     \cline{2-6} 
     I  &  VGG16$\_$pSC2   & 91.2 & -- & 3.27 M & 77.78 \\
     \cline{2-6} 
     F  &  VGG16$\_$pSC1       & 89.5 & --  &  1.64 M & 88.89 \\
     \cline{2-6} 
     A  &  ResNet18$\_$pSC9    & \textbf{92.9} & -- & 11.17 M & ---\\
     \cline{2-6} 
     R  &  ResNet18$\_$pSC4    & 92.5 & -- & 5.06 M & 54.65 \\
     \cline{2-6} 
     10 &  ResNet18$\_$pSC2    & 91.1 & -- &  2.62 M& 76.56 \\
    \cline{2-6} 
        &  ResNet18$\_$pSC1    & 89.4 & -- &  1.39 M& 87.50 \\\hline\hline
        & VGG16$\_$pSC9   & \textbf{57.2} & 78.9  & 14.73 M & --- \\
    \cline{2-6}    
        & VGG16$\_$pSC4   & 56.1 & \textbf{79.1} & 6.55 M & 55.56  \\
    \cline{2-6}
    Tiny    & VGG16$\_$pSC2     & 54.2 & 78.2 & 3.27 M & 77.78 \\
    \cline{2-6}
    Image   & VGG16$\_$pSC1     & 52.5 & 76.7 & 1.64 M & 88.89 \\
    \cline{2-6}
    Net     & ResNet18$\_$pSC9  & \textbf{62.4} & \textbf{83.2} &  11.17 M & ---\\
    \cline{2-6}
        & ResNet18$\_$pSC4  & 61.7 & 83 & 5.06 M & 54.65 \\
    \cline{2-6}
        & ResNet18$\_$pSC2  & 60.2 & 82.7 & 2.62 M& 76.56 \\
    \cline{2-6}
        & ResNet18$\_$pSC1  & 59.0 & 82.2 & 1.39 M & 87.50\\\hline    
    \end{tabular}
    }
    \end{minipage}
  \label{tab:vgg_res_pSC_cifar_tiny}
\end{table}

\begin{table}[h]
  \centering
   \caption{Test accuracy of different variants of periodic sparse kernel based VGG16 and ResNet18 on CIFAR-10 and Tiny ImageNet. The baseline architectures of Table \ref{tab:vgg_res_pSC_cifar_tiny} are used as the reference for calculating the reduction in parameters.}
   \begin{minipage}{\columnwidth}
    \resizebox{\columnwidth}{!}{
    \begin{tabular}{|c|c|c|c|c|c|c|}
    \hline
    Data & Model & (KVS, $P$) & Top 1  & Top 5 & Parameters & Parameter   \\
    set &  &    & acc $(\%)$ & acc $(\%)$ &  & reduction (\%)\\ \hline
     C  &   VGG16$\_$PS4\_P4 & (4, 4) & \textbf{91.7} & -- & 6.55 M& 55.56 \\
    \cline{2-7}    
     I  &   VGG16$\_$PS2\_P6 & (6, 6) & 90.6 & -- &  3.27 M& 77.78 \\
     \cline{2-7}
    F   &   VGG16$\_$PS1\_P9 & (9, 9)   & 87.9 & -- & 1.64 M& 88.89 \\
    \cline{2-7}
    A   &   ResNet18$\_$PS4\_P4 & (4, 4) & \textbf{92.9} & -- & 5.06 M& 54.65 \\
    \cline{2-7}
    R   &   ResNet18$\_$PS2\_P6 & (6, 6) & 91.5 & --  & 2.62 M& 76.56 \\
    \cline{2-7}
    10  &   ResNet18$\_$PS1\_P9 & (9, 9)   & 89.6 & -- & 1.39 M& 87.50 \\\hline\hline
            &   VGG16$\_$PS4\_P4 & (4, 4) & \textbf{56.9} & \textbf{79.9} & 6.55 M& 55.56 \\
    \cline{2-7}
    Tiny    &   VGG16$\_$PS2\_P6 & (6, 6) & 53.9 & 77.8 & 3.27 M& 77.78 \\
    \cline{2-7}
    Image   &   VGG16$\_$PS1\_P9 & (9, 9)   & 51.8 & 76.7 & 1.64 M& 88.89 \\
    \cline{2-7}
    Net     &   ResNet18$\_$PS4\_P4 & (4, 4) & \textbf{61.9} & \textbf{83} & 5.06 M& 54.65 \\
    \cline{2-7}
            &   ResNet18$\_$PS2\_P6 & (6, 6) & 60.7 & 82.9 & 2.62 M& 76.56 \\
    \cline{2-7}
            &   ResNet18$\_$PS1\_P9 & (9, 9)   & 58.9 & 81.8 & 1.39 M& 87.50 \\\hline
    \end{tabular}
    }
    \end{minipage}
  \label{tab:vgg_res_PS_P_cifar_tiny}
\end{table}

\subsection{Results for pSConv with Periodicity}
\label{subsec:preiodic_sparse}

The storage and energy advantage associated with periodically repeating kernels with some specific set of kernel variants, analysed in Section \ref{sec:evalofmerit}, motivated us to evaluate its performance in terms of test accuracy.  We leveraged the observation provided by \cite{ma2019pconv} and kept the KVS $= P$ small for different KSS based architectures. In particular, as KSS of 4 covers more kernel entries per variant, we chose a corresponding P = KVS = 4 and covered all possible kernel entries of the $3 \times 3$ kernels. For similar reasons, we chose larger KVS for KSS of 2 and 1, respectively (6 and 9, respectively). 
We selected kernel variants as described in Section \ref{subsec:psconv}. 
Fig. \ref{fig:Acc_error_cifar_tiny_res_vgg} shows the learning curves
for CIFAR-10 and Tiny ImageNet datasets with different variants of VGG16 and ResNet18 models with KSS of 1.\footnote{We saw similar trends with KSS of 2 and 4 in VGG16 and ResNet18, and so did not show in separate plots for brevity's sake.} It is clear that the sparse variants learn at similar rates as the corresponding baselines.

\begin{figure*}[t]
    \centering
        \subfloat[\label{subfig:scatter_plot_a}]{\includegraphics[width=6.0cm]{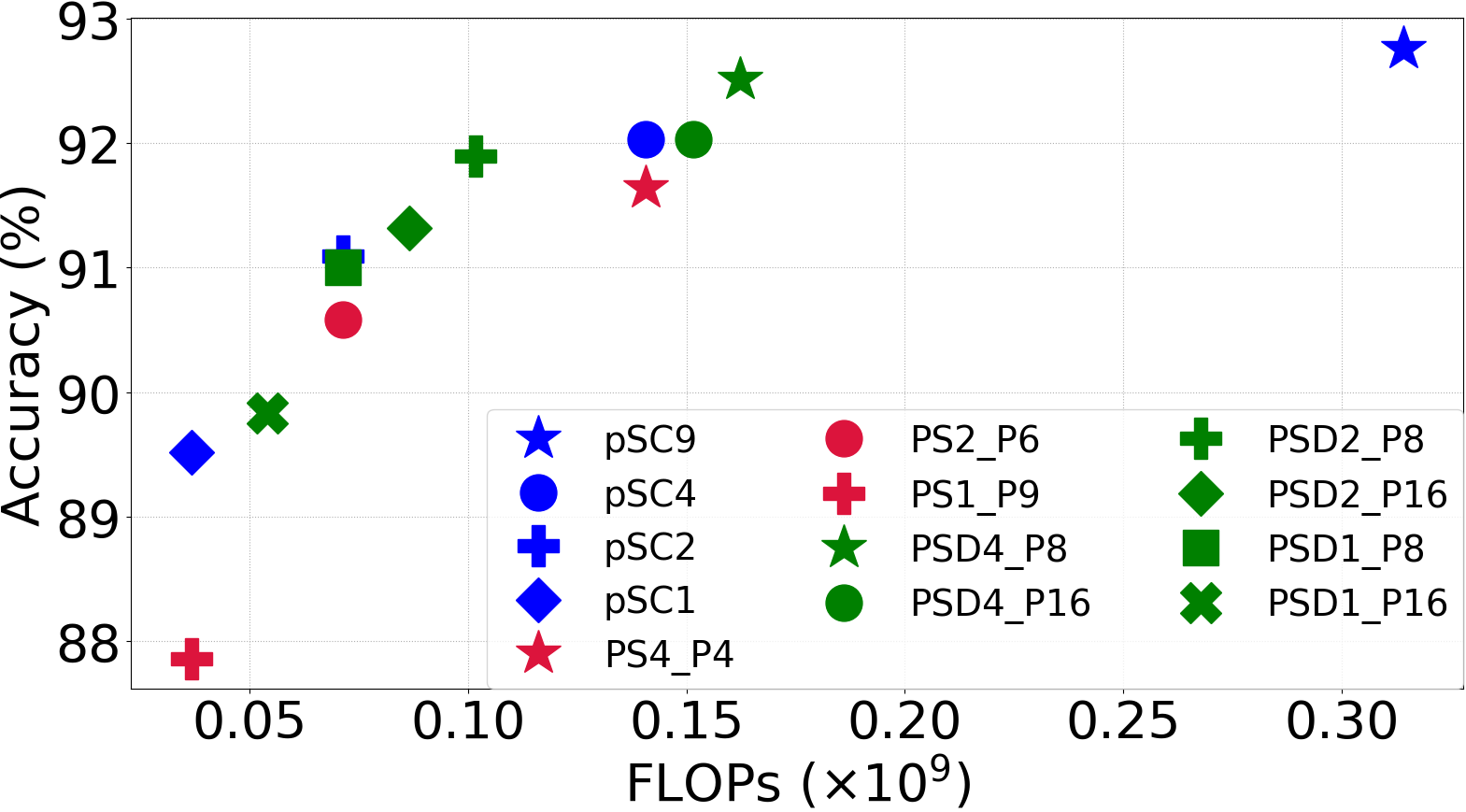}}\hspace*{1.5cm}
        \subfloat[\label{subfig:scatter_plot_b}]{\includegraphics[width=6.0cm]{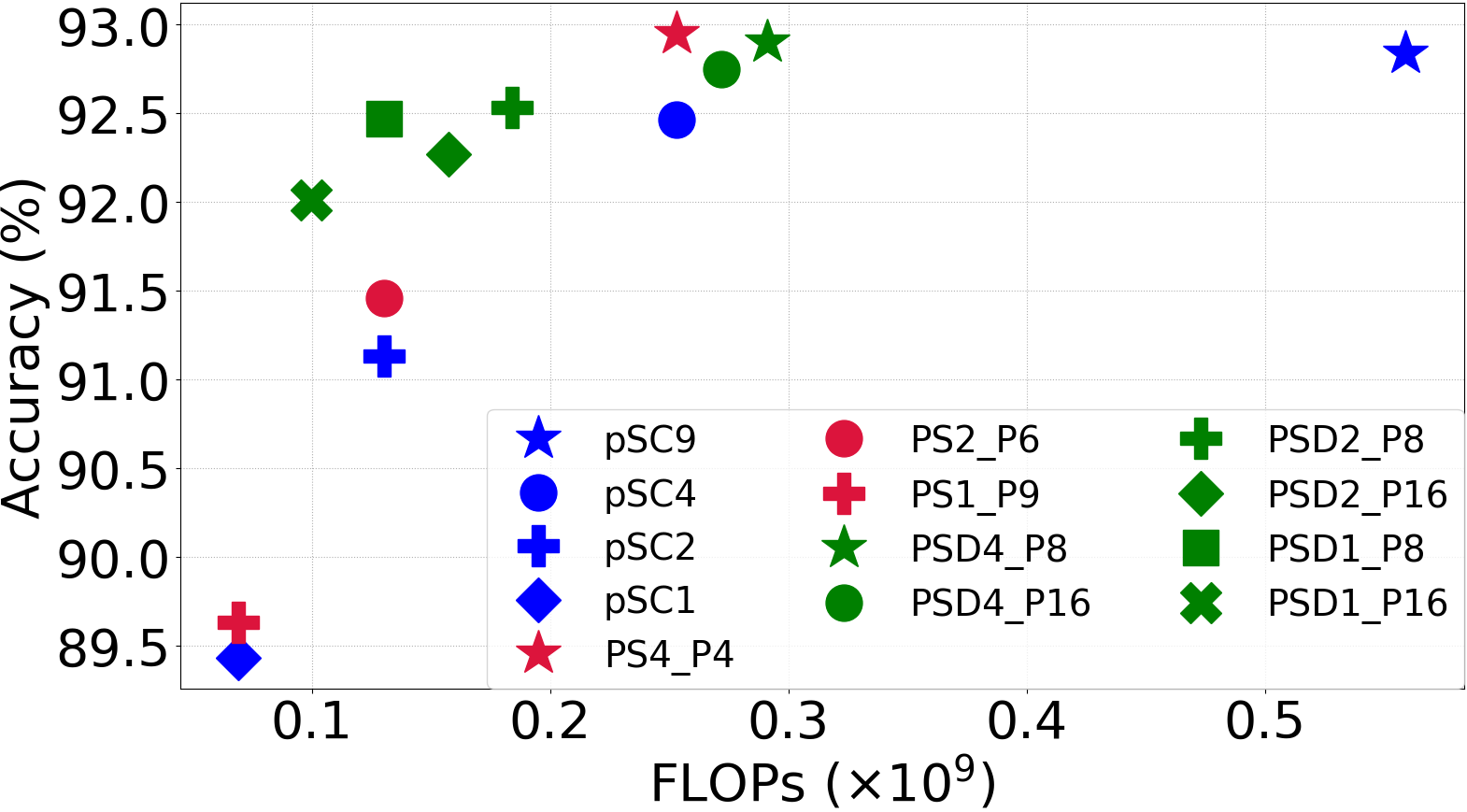}} \\
        \subfloat[\label{subfig:scatter_plot_c}]{\includegraphics[width=6.0cm]{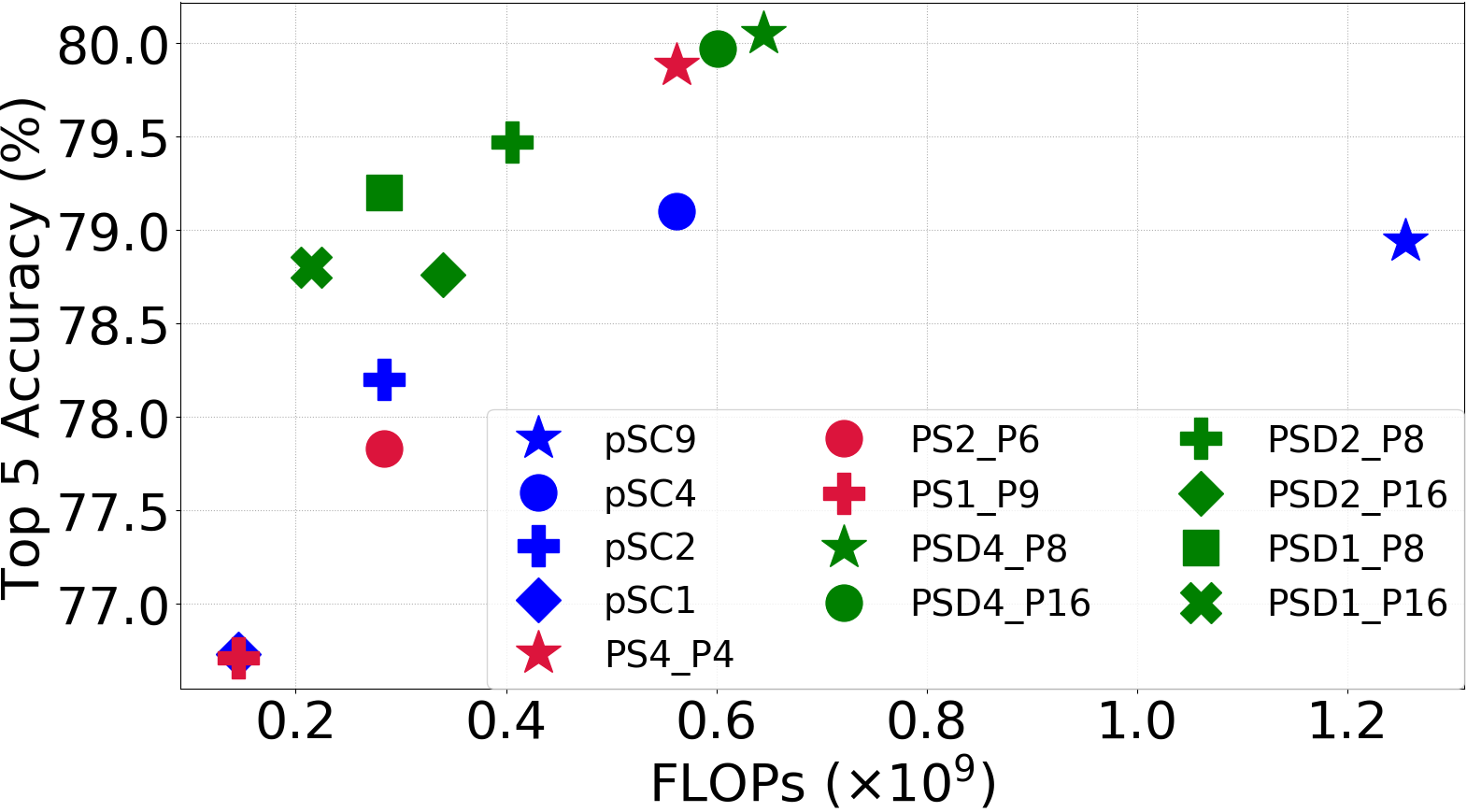}}\hspace*{1.5cm}
        \subfloat[\label{subfig:scatter_plot_d}]{\includegraphics[width=6.0cm]{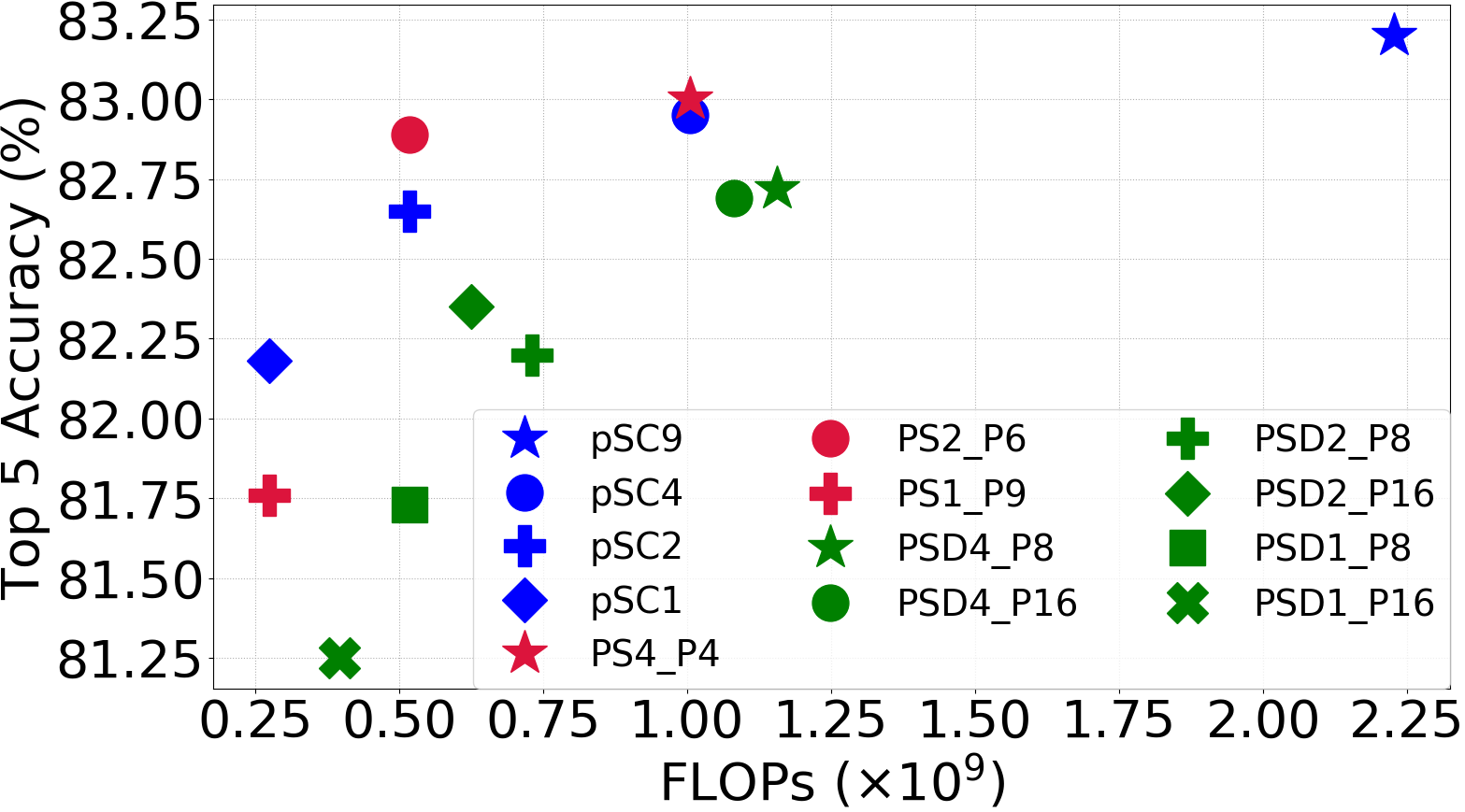}}\vfill
    \caption[Optional caption for list of figures]{Test accuracy vs. FLOPs count plots for different datasets on different architectures: CIFAR-10 on \subref{subfig:scatter_plot_a}  VGG16, \subref{subfig:scatter_plot_b} ResNet18 variants; Tiny ImageNet on \subref{subfig:scatter_plot_c} VGG16, \subref{subfig:scatter_plot_d} ResNet18 variants.}
    \label{fig:acc_vs_flops_plots}
\end{figure*}

Table \ref{tab:vgg_res_PS_P_cifar_tiny} shows the impact of an added periodicity constraint on test accuracy with our proposed variants.
Note that because of the overhead of storing auxiliary vectors, the overall storage reduction is smaller than the ones reported in Table~\ref{tab:vgg_res_PS_P_cifar_tiny}. 
For example, for VGG16\_PS4\_P4, the reduction in the number of parameters is 55.6\%, but including the storage of the auxiliary vectors in  $\mathrm{CSR_4}$ format, the reduction is  approximately 44.6\%.  If $\mathrm{CSR}$ format is used, the reduction in overall storage requirements, relative to the baseline is approximately 25\%. 
%

\subsection{Results for Boosting}
\label{subsec:boost}

The results without and with periodically repeating sparse kernel patterns discussed in Sections \ref{subsec:psconv} and \ref{subsec:preiodic_sparse}, respectively, show considerable performance degradation at low KSS values such as 1. 
This section presents the performance of the network models with the proposed boosting method in which we periodically incorporate FC kernels  ($k \times k$) in the 3D filter.\footnote{\rev{In this paper, we focus on results with one FC kernel per period, i.e., $\eta$ = 1. However, we also evaluated performance with larger values of $\eta$. For example, $\eta$ = 2 for $P$ = 16, yields similar accuracy as $\eta$ = 1 for $P$ = 8. Both models have similar parameter counts but the latter has significantly lower storage costs, suggesting restricting our model space 
to have $\eta = 1$ is reasonable.}}

\rev{To evaluate the value of boosting, we measure its impact when periodicity $P$ is set to 8 and 16 as well as when applied to the non-boosting configurations used in Table \ref{tab:vgg_res_PS_P_cifar_tiny}.} We tested the same sparse kernel variants as those 
used in Section \ref{subsec:preiodic_sparse}. Thus, when the
number of unique variants are less than $P$, we randomly chose some of the sparse kernel variants to repeat before placing the FC kernels. However, for simulation of aaa\_PSD1\_P8 models we randomly choose 7 of 9 unique sparse kernel variants. Note that
because each period will now contain one FC kernel, the proposed criteria of covering all kernel entries within a period is automatically satisfied. 

\begin{table}[h!]
  \centering
   \caption{Test accuracy of different variants of VGG16, and ResNet18 on CIFAR-10 with periodic sparse kernels boosted through insertion of periodic FC kernels.}
   \begin{minipage}{\columnwidth}
   \resizebox{\columnwidth}{!}{
    \begin{tabular}{|c|c|c|c|c|c|}
    \hline
     Model & (KVS, $P$) & Test & Improvement & Parameters & Parameter   \\
    & & acc $(\%)$ & over periodic & & reduction (\%)\\ \hline
     VGG16$\_$PSD4\_P8 & (5, 8) & 92.5  & +0.87 & 7.57 M & 48.61 \\\hline
     VGG16$\_$PSD4\_P16 & (5, 16) & 92.0 & +0.39 & 7.06  M & 52.1 \\\hline
     VGG16$\_$PSD2\_P8 & (7, 8) & 91.9 & +1.32 & 4.71 M & 68.1 \\\hline
    VGG16$\_$PSD2\_P16 & (7, 16) & 91.3  & +0.74 & 3.99 M & 72.92 \\\hline
    VGG16$\_$PSD1\_P8 & (8, 8) & 91 & +3.14 & 3.27 M & 77.78 \\\hline
     VGG16$\_$PSD1\_P16 & (10, 16) & 89.8 & +1.97 & 2.46 M & 83.33 \\\hline
     \rev{VGG16$\_$PSD4\_P4} & \rev{(4, 4)} & \rev{92.4} & \rev{+0.77} & \rev{8.59 M}  & \rev{41.67} \\\hline
     \rev{VGG16$\_$PSD2\_P6} & \rev{(6, 6)} & \rev{92} & \rev{+1.42} & \rev{5.18 M} & \rev{64.81} \\\hline
     \rev{VGG16$\_$PSD1\_P9} & \rev{(9, 9)} & \rev{91.05} & \rev{+3.22} & \rev{3.09 M} & \rev{79} \\\hline\hline
     ResNet18$\_$PSD4\_P8 & (5, 8) & 92.9 & +0.00 & 5.82 M & 47.83 \\\hline
     ResNet18$\_$PSD4\_P16 & (5, 16) & 92.8 & -0.15 & 5.43 M & 51.26 \\\hline
     ResNet18$\_$PSD2\_P8 & (7, 8) & 92.5 & +1.09 & 3.68 M & 67 \\\hline
     ResNet18$\_$PSD2\_P16 & (7, 16) & 92.3 & +0.81 & 3.15 M & 71.78 \\\hline
     ResNet18$\_$PSD1\_P8 & (8, 8) & 92.5 & +2.84 & 2.62 M & 76.56 \\\hline
     ResNet18$\_$PSD1\_P16 & (10, 16) & 92.0 & +2.4 & 2.01 M & 82.02 \\\hline
     \rev{ResNet18$\_$PSD4\_P4} & \rev{(4, 4)} & \rev{\textbf{93.0}} & +0.1 & \rev{6.58 M} & \rev{41} \\\hline
     \rev{ResNet18$\_$PSD2\_P6} & \rev{(6, 6)} & \rev{92.4} & \rev{+0.9} & \rev{4.04 M} & \rev{63.8} \\\hline
     \rev{ResNet18$\_$PSD1\_P9} & \rev{(9, 9)} & \rev{92.2} & \rev{+2.6} & \rev{2.48 M} & \rev{77.77} \\\hline
    \end{tabular}
    }
    \end{minipage}
  \label{tab:vgg_res_PSD_P_cifar}
\end{table}

\begin{table}[h]
  \centering
   \caption{Test accuracy of different variants of VGG16, and ResNet18 on Tiny ImageNet with periodic sparse kernels boosted with periodic FC kernels.}
   \begin{minipage}{\columnwidth}
    \resizebox{\columnwidth}{!}{
    \begin{tabular}{|c|c|c|c|c|c|}
    \hline
    Model & (KVS, $P$) & Top 1 & Improvement & Parameters & Parameter  \\
    & & acc $(\%)$&  over periodic &  & reduction  (\%) \\ \hline
    VGG16$\_$PSD4\_P8 & (5, 8) & \textbf{57.3} & +0.35 & 7.57 M & 48.61 \\\hline
    VGG16$\_$PSD4\_P16 & (5, 16) & 56.9 & +0.0 & 7.06 M & 52.1 \\\hline
    VGG16$\_$PSD2\_P8 & (7, 8) & 55.9 & +1.95 & 4.71 M & 68.1 \\\hline
    VGG16$\_$PSD2\_P16 & (7, 16) & 55.5 & +1.55 & 3.99 M & 72.92 \\\hline
    VGG16$\_$PSD1\_P8 & (8, 8) & 55.3 & +3.55 & 3.27 M & 77.78 \\\hline
    VGG16$\_$PSD1\_P16 & (10, 16) & 55.1 & +3.3 & 2.46 M & 83.33 \\\hline
    \rev{VGG16$\_$PSD4\_P4} & \rev{(4, 4)} & \rev{57.3} & \rev{+0.35} & \rev{8.6 M} & \rev{41.67} \\\hline
    \rev{VGG16$\_$PSD2\_P6} & \rev{(6, 6)} & \rev{56.3} & \rev{+2.35} & \rev{5.18 M} & \rev{64.81} \\\hline
    \rev{VGG16$\_$PSD1\_P9}& \rev{(9, 9)} & \rev{55} & \rev{+3.2} & \rev{3.09 M} & \rev{79} \\\hline\hline
    ResNet18$\_$PSD4\_P8 & (5, 8) & \textbf{61.8} & -0.09 & 5.82 M & 47.83 \\\hline
    ResNet18$\_$PSD4\_P16 & (5, 16) & 61.7 & -0.23 & 5.43 M & 51.26 \\\hline
    ResNet18$\_$PSD2\_P8 & (7, 8) & 60.6 & -0.13 & 3.68 M & 67 \\\hline
    ResNet18$\_$PSD2\_P16 & (7, 16) & 60.2 & -0.48 & 3.15 M & 71.78 \\\hline
    ResNet18$\_$PSD1\_P8 & (8, 8) & 60.0 & +1.15 & 2.62 M & 76.56 \\\hline
    ResNet18$\_$PSD1\_P16 & (10, 16) & 59.0 & +0.15 & 2.01 M & 82.02 \\\hline
    \rev{ResNet18$\_$PSD4\_P4} & \rev{(4, 4)} & \rev{\textbf{62.9}} & \rev{+1.0} & \rev{6.58 M} & \rev{41} \\\hline
    \rev{ResNet18$\_$PSD2\_P6} & \rev{(6, 6)} & \rev{60.5} & \rev{-0.23} & \rev{4.04 M} & \rev{63.8} \\\hline
    \rev{ResNet18$\_$PSD1\_P9} & \rev{(9, 9)} & \rev{59.6} & \rev{+0.75} & \rev{2.48 M} & \rev{77.77} \\\hline
    \end{tabular}
    }
    \end{minipage}
  \label{tab:vgg_res_PSD_P_tiny}
\end{table}

Table \ref{tab:vgg_res_PSD_P_cifar} and \ref{tab:vgg_res_PSD_P_tiny} show the classification accuracy
improvement compared to their sparse periodic counterparts
and parameter count reduction compared to the 
corresponding baseline models. 
The results show that boosting yields an improvement of up to 3.2\% (3.6\%) in classification accuracy for CIFAR-10 (Tiny ImageNet). With sparse KSS of 4, the average performance improvement compared to periodic sparse models is $\mathord{\sim} 0.3\%$. This is quite intuitive as the potential improvement is lower when KSS is high. However, for low KSS the average improvement is $\mathord{\sim} 2.3\%$. For example, ResNet18 with KSS of 1 and repeating FC kernels with a period of 8 on CIFAR-10 provides an accuracy degradation of only $\mathord{\sim}0.4\%$ compared to the baseline, which was earlier $\mathord{\sim}3.3\%$ without the FC kernels inserted. This motivates the use of boosted pre-defined kernels that are very sparse. We observed similar trends with Tiny ImageNet as well. 
The relative cost of the increase in parameters due to boosting is low and, as the periodicity of the fully connected kernel placement increases, it becomes negligible.
Fig. \ref{fig:acc_vs_flops_plots} shows the accuracy vs. FLOPs\footnote{We consider FLOPs associated with only the convolution layers because they generally represent the vast majority of FLOPs.} relation for different architecture variants. Models whose points lie towards the top-left have better accuracy with fewer FLOPs. In particular, for VGG16 and ResNet18 variants on CIFAR-10 and VGG16 variants on Tiny ImageNet, boosting performs consistently well, whereas, as we can see from Fig. \ref{fig:acc_vs_flops_plots} (d), boosting is not as beneficial for Tiny ImageNet on ResNet18. In general, we see that, with modest computation overhead, boosting consistently improves accuracy for models with extremely low KSS and maintains high accuracy otherwise.

It is important to emphasize that the overall parameter overhead is a function of both periodicity and KSS, as exemplified by the four sparse models described in Table \ref{tab:actual_param_reduc} analyzed using 
the storage requirement formulas in Table \ref{tab:smsf_comparison}.
Comparing models 1 and 2, which have  the same sparse KSS,  shows the impact of periodicity; as does comparing models 3 and 4.  In contrast, comparing  models 1 and 3 shows the impact of KSS for fixed periodicity; as does comparing models 2 and 4.  
The last two columns of the table represent the parameter counts normalized with respect to the baseline model.
Averaging across the four examples, the table shows that $CSR_{P}$ reduces the overall parameter count compared to $CSR$, including the sparse
matrix representation, by $22\%$.
Perhaps more importantly, the results show that the $CSR_P$ format can reduce the overall parameter count by as much as $70\%$ compared to the baseline model.

\rev{To better evaluate the space and choice of KVs, we generated  model variants with six different random seeds. 
We tested VGG16 and ResNet18 models with $KSS$ of 4 and 2 to classify CIFAR-10 and Tiny ImageNet. We observed differences of less than $1\%$ between the minimum and maximum classification accuracy across the different seeds.
In particular, for ResNet18\_PSD2\_P8 and ResNet18\_PSD4\_P8 the gaps between minimum and maximum accuracy are 0.55\% and 0.44\%, respectively, averaged over the two datasets. For VGG16\_PSD2\_P8 and VGG16\_PSD4\_P8 these values are 0.65\% and 0.65\%, respectively.}  
\begin{table}[h]
 \centering
 \caption{Parameters reduction and corresponding normalized storage requirement including indexing overhead for four VGG16 variants with both $CSR_{P}$ and $CSR$ format of compressed storage.}
  \begin{minipage}{\columnwidth}
   \resizebox{\columnwidth}{!}{
   \begin{tabular}{|c|c|c|c|c|}
    \hline
    No. &Model & Model param.  & Normalized param.   & Normalized param.\\
    & & reduction (\%) &  count, using $CSR_P$ & count, using $CSR$ \\\hline
    1 & VGG16\_PSD4\_P8 & 48.61 & 0.66 & 0.85 \\\hline
    2 & VGG16\_PSD4\_P16 & 52.10 & 0.69 & 0.81 \\\hline
    3 & VGG16\_PSD1\_P8 & 77.78 & 0.34 & 0.42 \\\hline
    4 & VGG16\_PSD1\_P16 & 83.33 & 0.30 & 0.35 \\\hline
    
    \end{tabular}
    }
   \end{minipage}
 \label{tab:actual_param_reduc}
\end{table}

Lastly, to demonstrate boosting has general benefits, Table \ref{tab:vgg_res_pSC_P_tiny} shows the results of boosting with Tiny ImageNet\footnote{For the CIFAR-10 dataset we obtained similar results, with ResNet18\_pSC4\_P8 exceeding the baseline performance with an average test accuracy of 92.95\%.} when the FC kernels are placed periodically, with period $P_D$, in between sparse kernels {\em with no pre-defined KVS or kernel variants} (as described in Section \ref{subsec:psconv}). 
Note, as with the networks described in Section \ref{subsec:psconv}, the lack of structure makes these models have higher indexing overhead compared to the periodic models analyzed above.



\begin{table}[h]
  \centering
  \caption{Test accuracyn of boosting as a general method to improve accuracy. Dataset used here is Tiny ImageNet.}
   \begin{minipage}{\columnwidth}
    \resizebox{\columnwidth}{!}{
    \begin{tabular}{|c|c|c|c|c|}
    \hline
    Model & (KVS, $P_D$) & Top 1 & Parameters & Parameter   \\
    & & acc $(\%)$ &  & reduction (\%)\\ \hline
    VGG16$\_$pSC4\_P8 & (--, 8) & \textbf{56.6} & 7.57 M & 48.61 \\\hline
    VGG16$\_$pSC4\_P16 & (--, 16) & 56.2 & 7.06 M & 52.1 \\\hline
    VGG16$\_$pSC2\_P8 & (--, 8) & 56.6 & 4.71 M & 68.1 \\\hline
    VGG16$\_$pSC2\_P16 & (--, 16) & 56.4 & 3.99 M & 72.92 \\\hline
    VGG16$\_$pSC1\_P8 & (--, 8) & 55.5 & 3.27 M & 77.78 \\\hline
    VGG16$\_$pSC1\_P16 & (--, 16) & 54.8 & 2.46 M & 83.33 \\\hline\hline
    ResNet18$\_$pSC4\_P8 & (--, 8) & 61.8  & 5.82 M & 47.83 \\\hline
    ResNet18$\_$pSC4\_P16 & (--, 16) & \textbf{62.3} & 5.43 M & 51.26 \\\hline
    ResNet18$\_$pSC2\_P8 & (--, 8) & 61.3 & 3.68 M & 67 \\\hline
    ResNet18$\_$pSC2\_P16 & (--, 16) & 60.5 & 3.15 M & 71.78 \\\hline
    ResNet18$\_$pSC1\_P8 & (--, 8) & 59.8 & 2.62 M & 76.56 \\\hline
    ResNet18$\_$pSC1\_P16 & (--, 16) & 59.2 & 2.01 M & 82.02 \\\hline
    \end{tabular}
    }
    \end{minipage}
  \label{tab:vgg_res_pSC_P_tiny}
\end{table}

\subsection{Performance Comparison with \rev{ShuffleNet and} MobileNetV2}

\begin{figure*}[h!]
\includegraphics[width=0.8\linewidth]{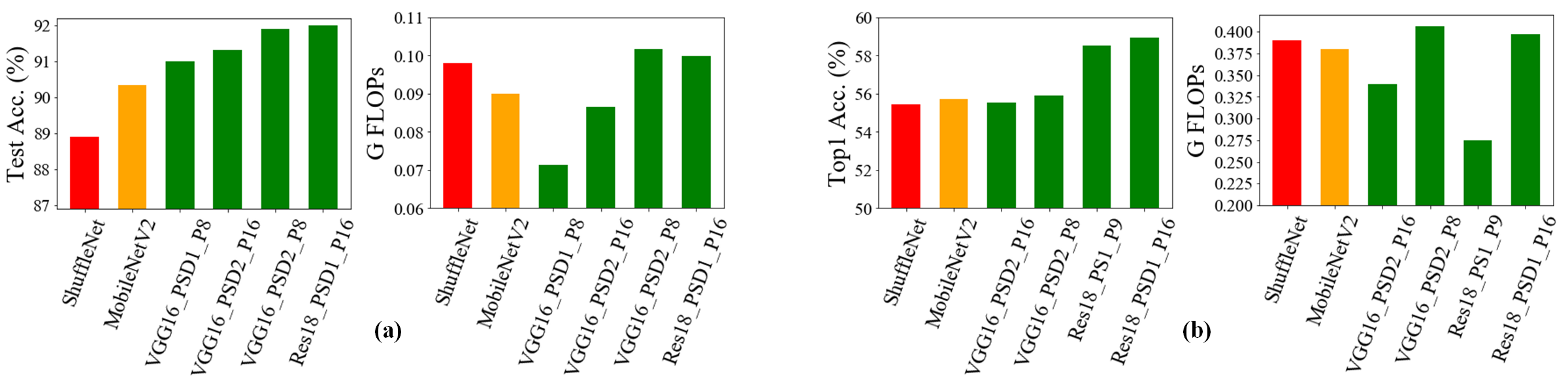}
\centering
\caption{\rev{Performance comparison of our proposed architectures that have similar or fewer FLOPs than ShuffleNet and MobileNetV2 with comparable or better classification accuracy on (a) CIFAR-10 and (b) Tiny ImageNet.} }
\label{fig:compare_mob_cifar_tiny}
\end{figure*}

\rev{Because ShuffleNet \cite{zhang2018shufflenet} and MobileNetV2 \cite{sandler2018mobilenetv2} are two widely-accepted low-complexity CNN architectures, we compared them with our proposed pre-defined periodic sparse models that have similar or fewer FLOPs}.\footnote{Note that we kept the hyperparameters for MobileNetV2 training the same as ResNet18 except the weight decay which was set to 0 as recommended by the original papers \cite{sandler2018mobilenetv2}.} 
\rev{In particular, Fig. \ref{fig:compare_mob_cifar_tiny}(a) shows that for CIFAR-10 the ResNet18\_PSD1\_P16 increases accuracy to 92\% compared to the baseline MobileNetV2 (ShuffleNet) accuracy of 90.3\% ($\mathord{\sim}89\%$).
Note that our obtained accuracies are also superior than reported in \cite{lawrence2019iotnet} and only around 1\% less than the accuracy reported in \cite{she2019scienet} which was trained for 180 additional epochs.
The pre-defined sparse CNN model VGG16\_PSD1\_P8 With 0.073 G FLOPs, has approximately $1.24\times$ ($1.34\times$) fewer computation complexity yet still outperforms MobileNetV2 (ShuffleNet) in terms of accuracy. For Tiny ImageNet, 
as shown in 
Fig.~\ref{fig:compare_mob_cifar_tiny}(b), our best classifying model provides an accuracy improvement of 3.2\% with only 4\% (2.6\%) increased complexity compared to MobileNetV2 (ShuffleNet).} 

\begin{figure}[h!]
\includegraphics[width=0.8\linewidth]{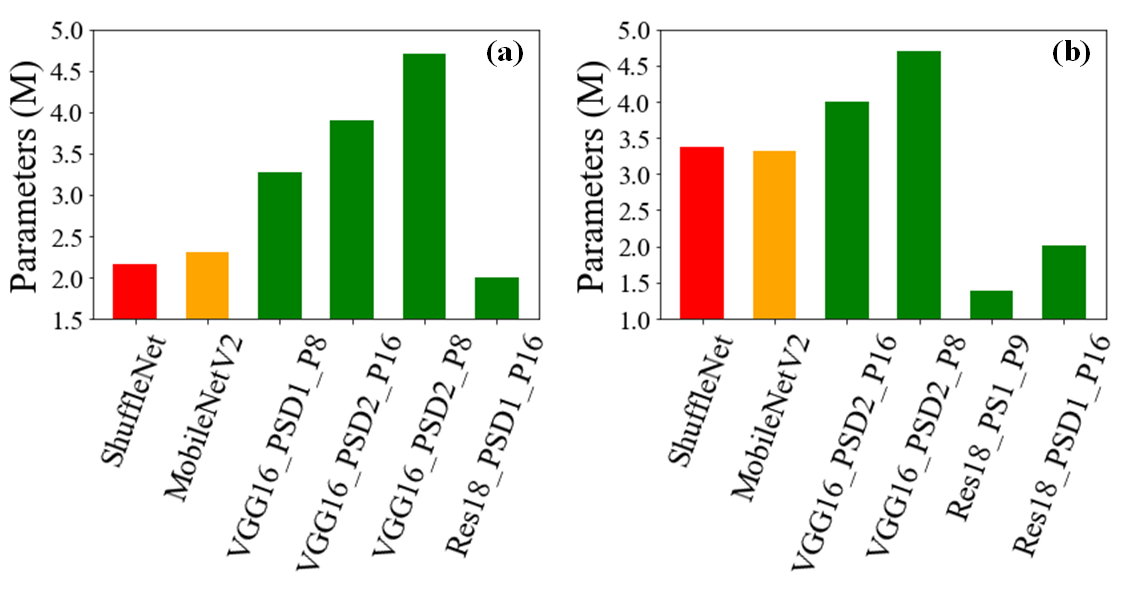}
\centering
\caption{\rev{Comparison of the number of model parameters of the network models described in Fig \ref{fig:compare_mob_cifar_tiny} for (a) CIFAR-10 and (b) Tiny ImageNet datasets.}}
\label{fig:mob_compare_params_cifar-tiny}
\end{figure}

\rev{Moreover, as we can see from Fig. \ref{fig:mob_compare_params_cifar-tiny}(a), and (b), with $2.42\times$ ($1.08\times$) fewer parameters our proposed models perform similar to ShuffleNet for Tiny ImageNet (CIFAR-10). Similarly,
the parameter requirement of our proposed models with similar accuracy as MobileNetV2 are $1.15 \times$, and $2.38 \times$ lower for CIFAR-10 and Tiny ImageNet, respectively.}\footnote{These values can be translated to the normalized parameter count with the help of the formulas in Table \ref{tab:smsf_comparison}.} 

\subsection{Performance Evaluation on Networks Models with Scaled Down Width}
\label{subsec:width_mult}

\begin{figure*}[h!]
\includegraphics[width=0.95\linewidth]{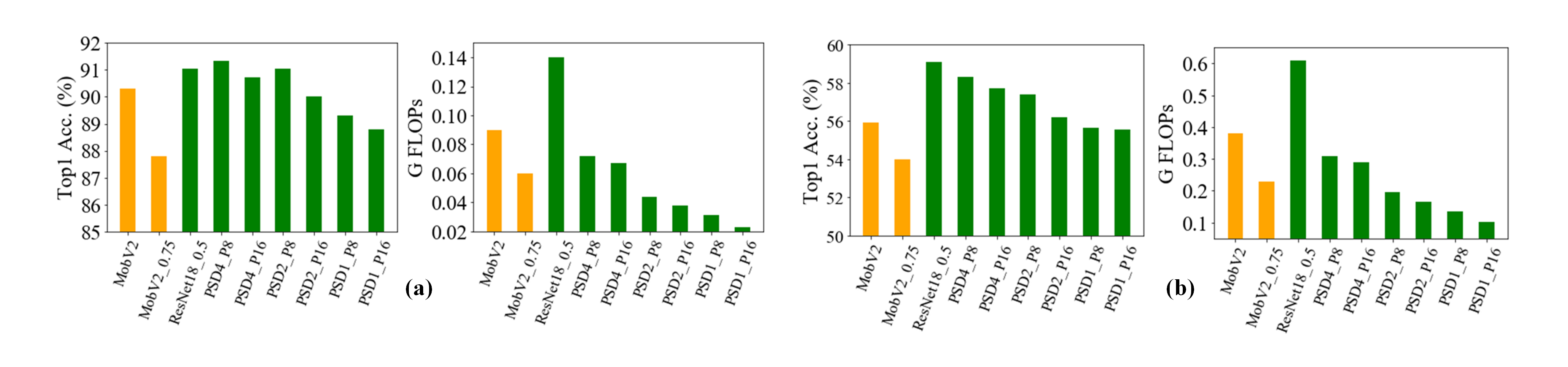}
\centering
\caption{Performance comparison in terms of test accuracy and FLOPs of different squeezed (width multiplier 0.5) ResNet18 variant models with  MobileNetV2 (MobV2) having width multiplier 1.0 and 0.75 on (a) CIFAR-10, and (b) Tiny ImageNet. }
\label{fig:compare_squeez_res_mob_cifar_tiny}
\end{figure*}

Squeezing the network layers, i.e. reducing the number of channels per 3D filter by a factor of $\alpha$ (< 1.0), popularly known as the width multiplier, is another simple technique to reduce the network's FLOPs and storage requirement \cite{howard2017mobilenets, iandola2016squeezenet, tan2019efficientnet}. To further establish the idea of the pre-defined periodic sparsity, we apply our proposed kernels in squeezed variant of the ResNet18 architecture with an $\alpha$ of 0.5. The important network model parameters of the squeezed variants of ResNet18 and MobileNetV2 models are described in Table \ref{tab:squeeze_arch}.  With the same hyperparameter settings as stated in Section \ref{subsec:data_arch_hyper}, the baseline accuracy for ResNet18 with $\alpha$=0.5 are 91.1\%, and 59.1\% for CIFAR-10, and Tiny ImageNet, respectively. We trained several variants of this squeezed model with KSS values of 4, 2, and 1, each with the fully connected kernel repeating after every 8 and 16 kernels. Fig. \ref{fig:compare_squeez_res_mob_cifar_tiny} shows our proposed variants of squeezed ResNet18 consistently outperforms both MobileNetV2\_0.75 and MobileNetV2 in classification accuracy, keeping the number of FLOPs similar or lower. In particular, Fig. \ref{fig:compare_squeez_res_mob_cifar_tiny} (a) shows that on CIFAR-10 dataset, to provide similar accuracy the squeezed ResNet18 with KSS of 2 and periodicity of 16 requires $2.36 \times$ fewer FLOPs compared to MobileNetV2. Also, the ResNet18 variant that requires the least number of FLOPs, provides $\mathord{\sim}1\%$ improved accuracy with $2.6\times$ fewer computations compared to MobileNetV2\_0.75. A similar trend is observed for Tiny ImageNet, as shown in Fig. \ref{fig:compare_squeez_res_mob_cifar_tiny}(b). Averaged over the two datasets, the proposed squeezed ResNet18 variants provides similar accuracy with  $2.42\times$, and $2.37\times$ fewer 
FLOPs compared to MobileNetV2\_0.75 and MobileNetV2, respectively. On the same datasets, when we constrain the number FLOPs to be similar, pre-defined periodic sparsity can provide an average accuracy improvement of $\mathord{\sim}3.16\%$ and $\mathord{\sim}2.48\%$, compared to MobileNetV2 with $\alpha$ of 0.75 and 1.0, respectively. The model parameter reduction factors are proportional to the computation reduction and as the ResNet18\_0.5 model has comparable parameters as MobileNetV2, advantage in storage for the sparse versions of ResNet18\_0.5 is quite clear, and thus not discussed in details for brevity's sake. 

\begin{table}[h]
 \centering
 \caption{CONV layer channel width parameters with different $\alpha$ values of the network models.}
  \begin{minipage}{\columnwidth}
   \resizebox{\columnwidth}{!}{
   \begin{tabular}{|c|c|c|}
    \hline
    Name & $\alpha$ &Convolution layer different channel sizes \\\hline
    ResNet18 & 1.0 & [64, 128, 256, 512] \\                      
                \hline
    ResNet18\_0.5 & 0.5 & [32, 64, 128, 256]\\\hline
    MobileNetV2 & 1.0 &  [16, 24, 32, 64, 96, 160, 320] \\ \hline
    MobileNetV2\_0.75 & 0.75 & [12, 18, 24, 48, 72, 120, 240] \\\hline
    \end{tabular}
    }
   \end{minipage}
 \label{tab:squeeze_arch}
\end{table}

\section{Conclusions}
\label{sec:conc}

This paper showed that with pre-defined sparsity in convolutional kernels the network models can achieve significant model parameter reduction during both training and inference without significant accuracy drops. However, managing sparsity requires matrix indexing overhead in terms of storage and energy efficiency. To address this shortcoming, we added periodicity to the sparsity, periodically using same sparse kernel patterns in the convolutional layers,   significantly reduce the indexing overhead.

Furthermore, to deal with the performance degradation due to pre-defined sparsity, we introduced a low-cost network architecture modification technique in which FC kernels are periodically inserted in between sparse kernels. 
Experimental results showed that, compared to the sparse-periodic variants, this boosting technique improves average classification accuracy by up to $\mathord{\sim}2.3\%$, averaged over two periodicity of 8, and 16 in ResNet18 and VGG16 architecture on CIFAR-10 and Tiny ImageNet.
We also demonstrated the merits of the proposed architectures with squeezed variants of ResNet18 (width multiplier < 1.0) and  have shown it to outperform MobileNetV2 by an average accuracy of $\mathord{\sim}2.8\%$ with similar FLOPs.  

Our future work includes exploring additional forms of compressed sparse representations and their hardware support. \rev{Lastly, we note that much of our findings are empirical in nature. Finding a more theoretical basis that can motivate and guide the use of periodic pre-defined sparsity in deep learning is also an important area of future work.}
\bibliographystyle{ieeetr}
\vspace{-2mm}
\bibliography{biblio}

\begin{IEEEbiography}[{\includegraphics[width=1in,height=1.25in,clip,keepaspectratio]{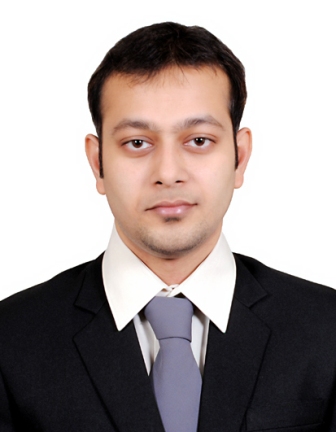}}]
{Souvik Kundu}
received his B.\,Tech degree in Electronics and Communication Engineering from West Bengal University of Technology in 2009 and M.\,Tech degree in Microelectronics and VLSI design from Indian Institute of Technology Kharagpur, India in 2015. He worked as R \& D Engineer II  at Synopsys India Pvt. Ltd. and as Digital Design Engineer at Texas Instruments India Pvt. Ltd. from 2015 to 2016 and from 2016 to 2017, respectively. He is currently working towards the Ph.D. degree in Electrical and Computer Engineering at the University of Southern California, Los Angeles, CA, USA. His research focuses on energy aware sparsity, model search, algorithm-hardware co-design of neural networks in machine learning.
\end{IEEEbiography}

\begin{IEEEbiography}[{\includegraphics[width=1in,height=1.25in,clip,keepaspectratio]{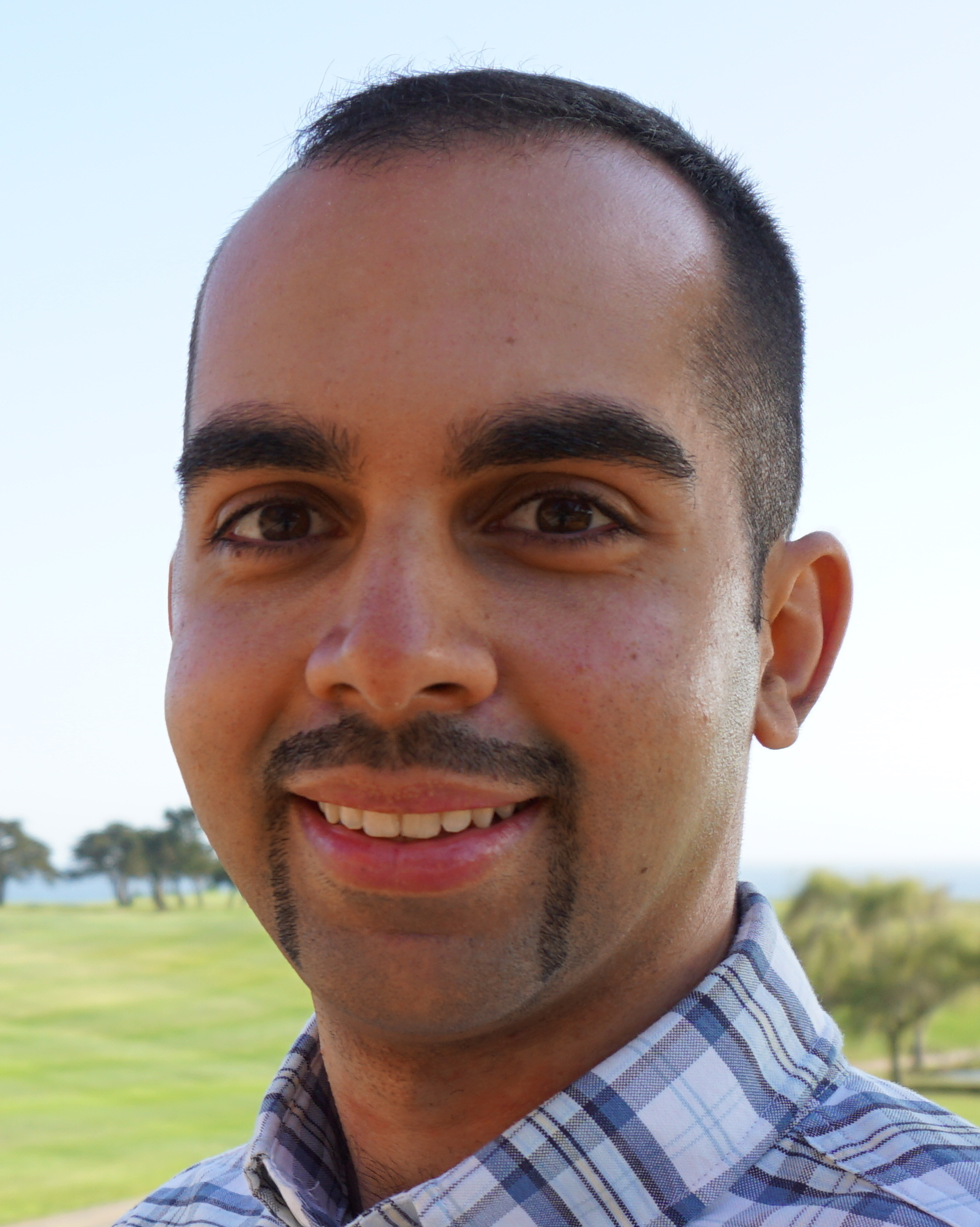}}]
{Mahdi Nazemi} received the B.S. degree in electrical engineering from the University of Tehran, Tehran, Iran, in 2014. He is currently pursuing the Ph.D. degree in electrical engineering with the University of Southern California, Los Angeles, CA, USA. He was an R\&D software engineering intern at the Cadence’s Genus Synthesis Solution group during the summer of 2016 and a software engineering intern at the Microsoft’s Artificial Intelligence and Research engineering group during the summer of 2018. His research focuses on algorithm/hardware co-design for the efficient processing of machine learning algorithms including deep neural networks. One of his recent publications in this area won the best paper award at the Asia and South Pacific Design Automation Conference in 2019. 
\end{IEEEbiography}
\begin{IEEEbiography}[{\includegraphics[width=1in,height=1.25in,clip,keepaspectratio]{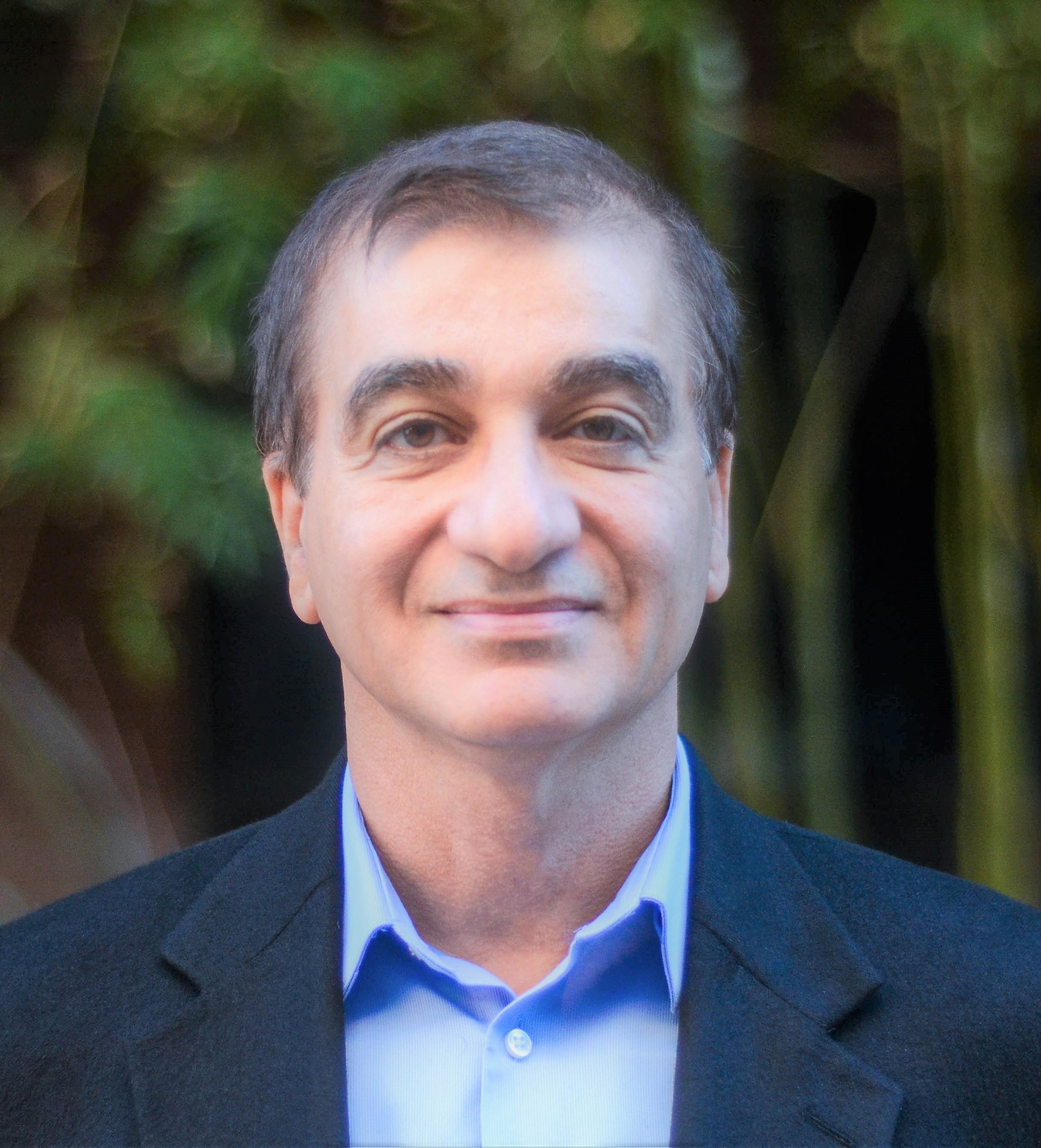}}]
{Massoud Pedram} obtained his B.S. degree in Electrical Engineering from Caltech in 1986 and Ph.D. in Electrical Engineering and Computer Sciences from the University of California, Berkeley in 1991. Subsequently, he joined the Department of Electrical Engineering of the University of Southern California where he currently holds the Charles Lee Powell Chair. His research interests include computer-aided design of VLSI circuits and systems, low power electronics, energy-efficient processing, electrical energy storage systems, power conversion and management ICs, quantum computing, and superconductive electronics. He has authored four books and more than 700 archival and conference papers. Dr. Pedram received the 2015 IEEE Circuits and Systems Society Charles A. Desoer Technical Achievement Award for his contributions to modeling and design of low power VLSI circuits and systems, and energy efficient computing and the 2017 USC Viterbi School of Engineering Senior Research Award. He also received the Third Most Cited Author Award at the 50th anniversary of the Design Automation Conf., Jun. 2013. Dr. Pedram is an IEEE Fellow. 
\end{IEEEbiography}
\begin{IEEEbiography}[{\includegraphics[width=1in,height=1.25in,clip,keepaspectratio]{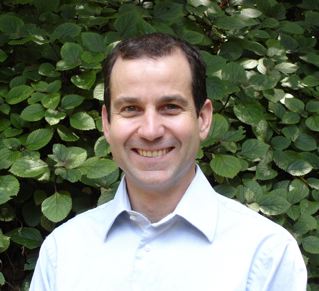}}]
{Keith M.~Chugg}(S'88-M'95-SM'06-F'10) received the B.S.~degree
(high distinction) in Engineering from Harvey Mudd College, Claremont, CA
in 1989 and Ph.D.~degree in Electrical Engineering  from
the University of Southern California (USC), Los Angeles, CA in 
1995.  Since 1996, he has been on the faculty of the Ming Hsieh Department of Electrical and Computer Engineering at USC, where he is currently a Professor.  His research interests are in the general areas of  signal processing, digital communications, machine learning, and associated efficient implementations.  He is a co-founder of TrellisWare
Technologies, Inc.,  where he serves as Chief Scientist.
\end{IEEEbiography}
\begin{IEEEbiography}[{\includegraphics[width=1in,height=1.25in,clip,keepaspectratio]{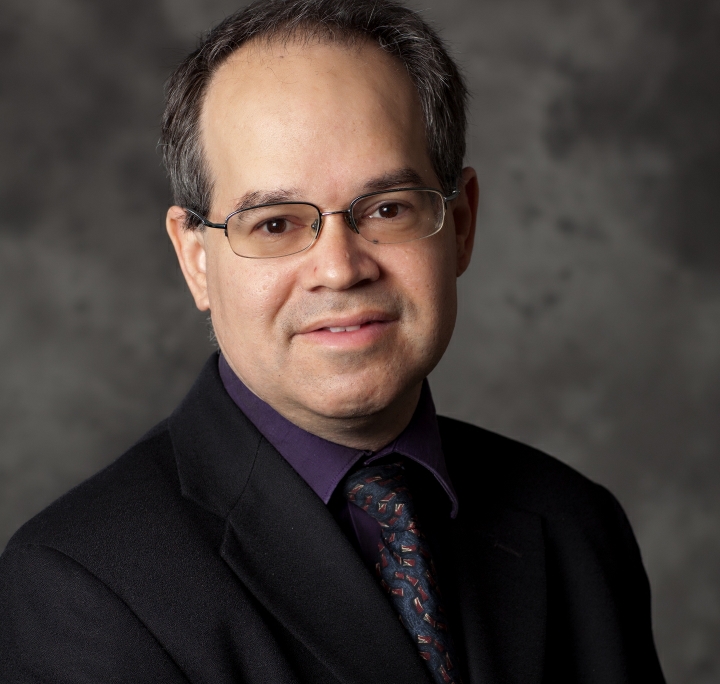}}]
{Peter A.~Beerel}(SM'08) received his B.S.E. degree in Electrical Engineering from Princeton University, Princeton, NJ, in 1989 and his M.S. and Ph.D. degrees in Electrical Engineering from Stanford University, Stanford, CA, in 1991 and 1994, respectively. He then joined the Ming Hsieh Department of Electrical and Computer Engineering at the University of Southern California where he is currently a professor and the Associate Chair of the Computer Engineering Division. He is also a Research Director at the Information Science Institute at USC. Previously, he co-founded TimeLess Design Automation to commercialize an asynchronous ASIC flow in 2008 and sold the company in 2010 to Fulcrum Microsystems which was bought by Intel in 2011. His interests include a variety of topics in computer-aided design, machine learning, hardware security, and asynchronous VLSI and the commercialization of these technologies. He is a Senior Member of the IEEE.
\end{IEEEbiography}

\end{document}